\documentclass[journal]{IEEEtran}

\usepackage{graphicx}
\usepackage{subfigure}
\usepackage{overpic}
\usepackage{amsmath}
\usepackage{flushend}
\usepackage{amssymb}
\usepackage{multirow}
\usepackage{makecell}
\usepackage{array}
\usepackage{stfloats}
\usepackage{algorithm}
\usepackage{algpseudocode}
\usepackage{amsmath}

\usepackage[normalem]{ulem}
\useunder{\uline}{\ul}{}

\usepackage{hyperref}

\hypersetup{
	colorlinks=true,
	linkcolor=cyan,
	urlcolor=red,
	citecolor=cyan,
}

\begin{document}

\title{Change Detection from Synthetic Aperture \\
Radar Images via Graph-Based Knowledge Supplement Network}
\author{Junjie Wang, Feng Gao \emph{Member, IEEE}, Junyu Dong, \emph{Member, IEEE},\\ 
 Shan Zhang, Qian Du, \emph{Fellow, IEEE}
\thanks{This work was supported in part by the National Key Research and Development Program of China under Grant 2018AAA0100602, in part by the Key Research and Development Program of Shandong Province under Grant 2019GHY112048, and in part by the Natural Science Foundation of Shandong Province under Grant ZR2019QD011. (\emph{Corresponding author: Feng Gao})}
\thanks{Junjie Wang, Feng Gao and Junyu Dong are with the School of Computer Science and Technology, Ocean University of China, Qingdao 266100, China.}
\thanks{Shan Zhang is with Meteorological Bureau of Tianjin Municipality, Tianjin 300000, China.}
\thanks{Qian Du is with the Department of Electrical and Computer Engineering, Mississippi State University, MS 39762, USA.}
}

\markboth{IEEE Journal Of Selected Topics In Applied Earth Observations And Remote Sensing}%
{Shell}

\maketitle
\begin{abstract}

Synthetic aperture radar (SAR) image change detection is a vital yet challenging task in the field of remote sensing image analysis. Most previous works adopt a self-supervised method which uses pseudo-labeled samples to guide subsequent training and testing. However, deep networks commonly require many high-quality samples for parameter optimization. The noise in pseudo-labels inevitably affects the final change detection performance. To solve the problem, we propose a \underline{G}raph-based \underline{K}nowledge \underline{S}upplement \underline{N}etwork (GKSNet). To be more specific, we extract discriminative information from the existing labeled dataset as additional knowledge, to suppress the adverse effects of noisy samples to some extent. Afterwards, we design a graph transfer module to distill contextual information attentively from the labeled dataset to the target dataset, which bridges feature correlation between datasets. To validate the proposed method, we conducted extensive experiments on four SAR datasets, which demonstrated the superiority of the proposed GKSNet as compared to several state-of-the-art baselines. Our codes are available at \verb'https://github.com/summitgao/SAR_CD_GKSNet' .

\end{abstract}


\IEEEpeerreviewmaketitle

\section{Introduction}

\IEEEPARstart{O}{wing} to the rapid development of earth observation programs, more multitemporal synthetic aperture radar (SAR) images are available, and they are captured over the same geographical area at different times. Since SAR images can be acquired under all-weather and all-time conditions, they have become the most important data source for change detection. SAR image change detection aims to accurately detect the changed information by analyzing two images captured at different times. It is of high practical value to a large number of applications, such as flood detection \cite{Giustarini12_tgrs}, disaster monitoring \cite{Brunner10}, urban planning \cite{Quan18}, land cover data monitoring \cite{Zanotta12}, and so on.

SAR images are inherently contaminated by multiplicative speckle noise, and this phenomenon makes the SAR image change detection a very challenging task. Therefore, it is essential to develop robust change detection techniques, which can cope with speckle noise. To solve the problem, researchers have devoted great efforts to put forward robust change detection methods. These methods can be broadly categorized into two main streams: supervised methods and unsupervised methods. A supervised method requires prior knowledge about land cover types or a large number of high-quality labeled samples \cite{Volpi13} \cite{Wang16}. In theory, supervised methods may offer better performance since many detailed descriptions of the changed region are provided. However, unsupervised methods are more popular since high-quality labeled samples are generally difficult to obtain in real applications \cite{Bruzzone00} \cite{Yetgin12}. Therefore, most existing methods are unsupervised methods. 

Unsupervised SAR image change detection methods are commonly composed of the following two steps: difference image (DI) generation and DI classification. In DI generation, the log-ratio \cite{Bazi06}, Gauss-ratio \cite{Hou14} and neighborhood-based ratio \cite{Gong12_grsl} operators are generally used, since these methods are considered robust to calibration errors \cite{Gao18}. In the DI classification step, clustering methods are widely employed to classify pixels into changed and unchanged classes, such as the fuzzy $c$-means (FCM) \cite{Mishra12}, $k$-means \cite{Celik09} and multiple kernel clustering \cite{Jia16}.

To enhance the performance of DI classification, researchers incorporate deep neural networks into the traditional unsupervised DI classification model. Hou et al. \cite{Hou17} presented a change detection method by combining deep features and saliency map computation using low-rank method. Zhan et al. \cite{Zhan17_grsl} proposed a deep siamese CNN model to extract discriminant features. Wang et al. \cite{wang19_getnet} proposed end-to-end 2D CNN framework for change detection. Mixed-affinity matrix was employed for DI analysis, and then CNNs were used to exploit the discriminant features. Du et al. \cite{du19_tgrs} proposed a slow feature analysis-based (SFA) method for change detection. Two deep networks were established to extract multitemporal features, and SFA was employed to extract the most invariant component of multitemporal features. Chen et al. \cite{chen20_tgrs} presented a deep siamese multiple-layers recurrent neural network (RNN) for change detection. Multiple-layers RNN was designed to handle the features extracted by CNN, which mapped features into a new space. In \cite{zhan20_tgrs}, a pretrained deep fully convolutional network is used for DI analysis, and on this basis, multiscale superpixel segmentation was employed to for robust change map generation. Zhao et al. \cite{zhao20_tgrs} proposed a metric learning-based generative adversarial network (GAN) for change detection, where metric learning was incorporated to enhance the stability of GAN model when the number of training samples was limited. Besides these methods, generative adversarial networks \cite{gong19_grsl} \cite{gong19_jstars}, fully convolution network \cite{liu20_jstars}, bipartite differential network \cite{liu20_tnnls}, and local restricted CNNs \cite{liu19_tnnls} are also employed to solve the problem of remote sensing image change detection.

Existing deep learning-based change detection methods commonly require many training samples to optimize network parameters. These samples are obtained by the self-learning strategy, which generates pseudo-labels from unlabeled DI pixels. Gong et al. \cite{Gong16_tnnls} assigned pseudo-labels to pixels in DI by an FCM-based joint classifier, then restricted Boltzmann machines (RBMs) were trained to generate the final change map. In \cite{Gao16_grsl}, a PCA-based neural network was introduced for SAR image change detection. PCA was employed as the cascaded filter for multitemporal feature analysis. Gao et al. \cite{gao19_cwnn} utilized convolutional-wavelet neural networks for SAR image change detection. Dual-tree complex wavelet transform is introduced for DI analysis, and the speckle noise can be suppressed effectively. In \cite{geng19_tgrs}, a nonnegative and Fisher-constrained autoencoder was designed to discover changed information from the DI. However, due to the limitations of clustering algorithms, these noisy pseudo-labeled samples contain error, and the error will be amplified during training \cite{li20_cvpr}. To solve the problem, we are dedicated to developing transfer learning-based change detection methods which can adapt additional knowledge from existing data with labels to new data without labels.

Some efforts have been made in transfer learning for change detection. Gao et al. \cite{gao19_grsl} presented a CNN-based transfer learning model for SAR image change detection. Liu et al. \cite{liu20_grsl} trained a U-Net from the source dataset and then transferred the pretrained model to the target dataset by minimizing a new designed loss function. Yang et al. \cite{yang19_tgrs} proposed a multitask transfer learning scheme for change detection. Two tasks were learned simultaneously: One for the source domain with labels, and the other for the unlabeled target data reconstruction. The aforementioned change detection methods use CNN or autoencoder model for knowledge transfer. We argue that if the knowledge from the source domain were distilled to target unlabeled data in more structured format, the change detection performance can be further improved.

Recently, there has been a surge of interest in graph-based methods. Graph reasoning has shown to have substantial practical merits for object detection \cite{xu19_cvpr} \cite{yan20_tip} \cite{liu20_cvpr}, image classification \cite{luo20_tgrs} \cite{chen19_cvpr1} \cite{chen19_cvpr2}, semantic segmentation \cite{zhang19_iccv} \cite{zhang19_icme} and change detection \cite{wang21grsl} \cite{wang21igarss} \cite{qu21tgrs}. Graph neural networks are powerful tools that can perform relational inference through message passing. The domain knowledge modeled in a single graph can be transferred to other graphs. Therefore, the graph-based method is natural to be adopted in the transfer learning-based change detection task. However, there are two problems regarding the following aspects:

\emph{1) How to suppress noisy samples in the target dataset via graph-based model?} In the target dataset, the pseudo-labeled samples selected from the DI inevitably contain some errors. If the model is blindly confident of these incorrect samples, the error will be amplified during training.  \emph{2) How to transfer knowledge among datasets with different characteristics?} SAR images captured by different satellite sensors have disparate feature representations of ground objects. It is challenging to transfer knowledge between datasets acquired by different sensors directly.

To handle the above-mentioned problems, we propose a \underline{G}raph-based \underline{K}nowledge \underline{S}upplement \underline{Net}work (GKSNet) for SAR image change detection. On the one hand, the extracted image features from existing labeled dataset are projected into a graph. After message propagation via graph convolutions, the obtained features are more discriminative, and these features are employed as additional knowledge for the target dataset. By knowledge supplement, more reliable information is introduced, and the adverse effects of noisy samples can be suppressed to some extent. On the other hand, a graph transfer module is proposed to distill contextual information attentively from the labeled dataset to the target dataset as supplementary knowledge. The knowledge bridges feature correlation from different datasets.

In summary, the main contributions of this article are threefold:

\begin{enumerate}

\item We perform SAR image change detection via a graph-based knowledge supplement network. Existing SAR image change detection methods cannot handle well errors in pseudo-labeled samples. The proposed network can suppress adverse effects of noisy samples by adding discriminative information from a labeled dataset.

\item In order to better integrate the supplementary knowledge, we propose a graph transfer module. Through feature fusion, the model can exploit the common knowledge and bridge the feature correlation between different datasets. Then, evolved features can be obtained to improve change detection performance.

\item We conducted extensive experiments on five SAR datasets to validate the effectiveness of GKSNet and the superiority of evolved features. As a side contribution, we have released our codes to benefit other researchers.

\end{enumerate}

The remainder of this paper is organized as follows. Section II presents the details of the proposed GKSNet, including intra-graph reasoning and inter-graph fusion. Section III provides the experimental results together with the corresponding analysis and discussion. Finally, conclusions are drawn in Section IV.

\begin{figure*}
\centering
\begin{center}
\includegraphics [width=6.5in]{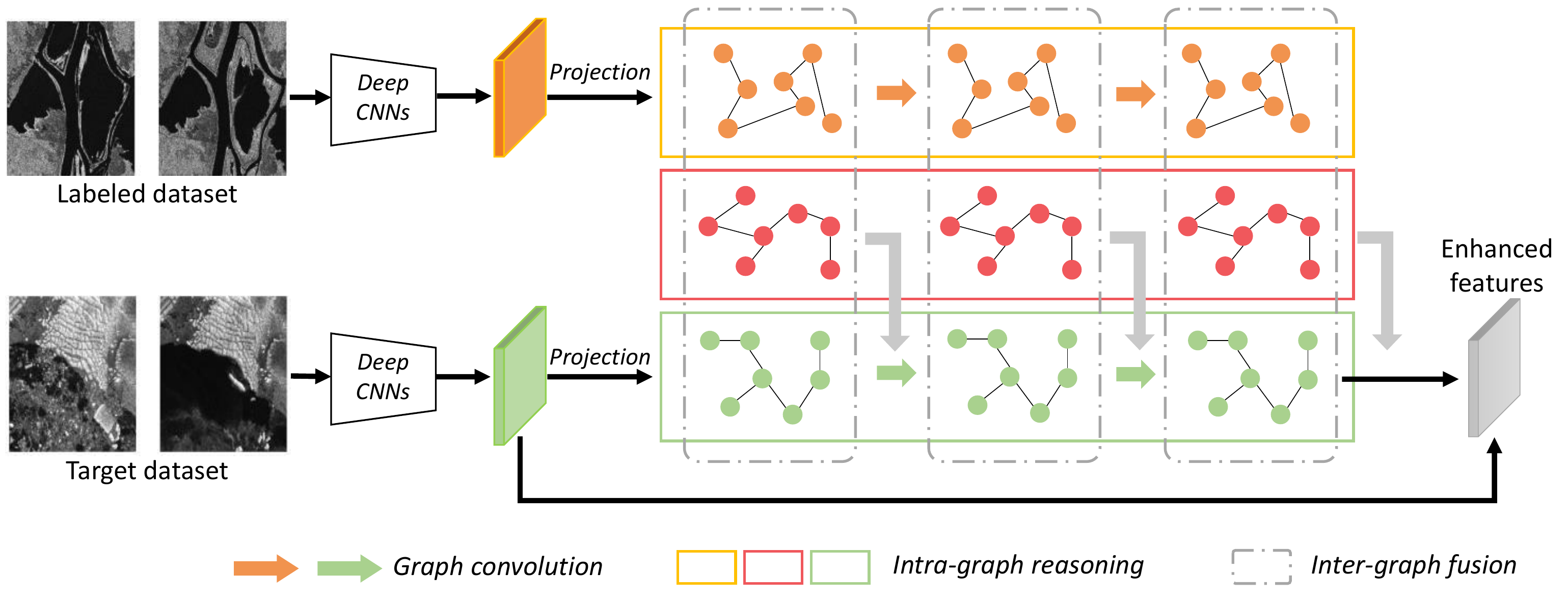}
\caption{Schematic illustration of the proposed GKSNet. The image features extracted by deep CNNs are projected into a graph representation. Then, the graph representations are transferred and fused via graph transfer module across different datasets. Finally, features from different graphs are fused via inter-graph fusion module. Through feature fusion, the model exploits the common knowledge and bridge the feature correlation from different datasets. The obtained evolved features are capable of improving the change detection performance.}
\label{structure}
\end{center}
\end{figure*}

\section{Methodology}

To alleviate the impact of noisy samples and enhance features, we aim at incorporating underlying knowledge from the labeled dataset via a graph-based network. Fig. \ref{structure} gives an overview of the proposed GSKNet. The proposed model can be embedded in any CNN-based classification model by enhancing its original convolution features vis graph transfer learning. Firstly, features extracted by CNNs are projected into a graph, and then the compact graph representation is learnt and propagated via intra-graph reasoning. Further, we transfer the graph representations using an inter-graph fusion module across different datasets.

\subsection{Preclassification and Reliable Samples Selection}

Given a pair of SAR images $I_1$ and $I_2$ that are captured over the same geographical area where an event of change happens, we aim to generate a binary change map $I_{cm}$: $I_{cm}(i,j)\in\{0, 1\}$, where 1 denotes that the position $(i,j)$ is changed, and 0 denotes that the position $(i, j)$ is unchanged. To this end, we need to create an initial change map with pseudo-labels via pre-classification. 

The first step of pre-classification is to generate a DI using the log-ratio operation. It is widely acknowledged that the log-ratio operator can reduce the influence of speckle noise, since it can transform multiplicative noise into additive noise and compress the range of values. After obtaining the DI, a pre-classification operation needs to be performed to obtain the pseudo-labels and training samples. Compared to other methods, a hierarchical clustering algorithm was proposed in \cite{Gao16_grsl}, which can better obtain enough representative samples for subsequent network training. Therefore, the hierarchical clustering algorithm \cite{Gao16_grsl} is employed to divide DI into three clusters $\{ \omega_c, \omega_u, \omega_i \}$, where $\omega_c$ and $\omega_u$ represent the changed and unchanged classes, respectively, and $\omega_i$ represents the uncertain class. Pixels belonging to $\omega_c$ have high probabilities to be changed, while pixels belonging to $\omega_u$ have high probabilities to be unchanged. Therefore, $\omega_c$ and $\omega_u$ are selected as reliable training samples for GKSNet training. Pixels from $\omega_i$ will be further classified by the GKSNet. To suppress the adverse effects of noisy samples, we reduce the number of training samples, which will be discussed in detail later.

The contextual information is critical for robust feature representation. Therefore, image patches centered at $\omega_c$ and $\omega_u$ are extracted from the original SAR images, and these patches are fed into the GKSNet as training samples. Let $R_k^1$ denote the image patch centered at pixel $k$ in $I_1$, and $R_k^2$ denote the corresponding image patch in $I_2$. The size of image patch is $r\times r$. Two patches are combined to form a training sample $R_k$ with the size of $r\times r\times 2$. It should be noted that $r$ is an important parameter, which will be discussed in Section III.

\subsection{Intra-Graph Reasoning}

Recently, many deep learning-based methods have been proposed for SAR image change detection. However, these methods commonly use a self-learning strategy to generate pseudo-labeled samples from unlabeled SAR data. However, these pseudo-labeled samples contain errors. These errors may be amplified during training \cite{jiang18_tgrs}. Therefore, ensuring robust feature representation while suppressing the errors in pseudo-labeled samples is the key to improving the change detection performance.

To address this issue, we propose a Graph-based Knowledge Supplement Network (GKSNet) which can extract the common knowledge existing in the labeled dataset as feature supplement to ensure robust parameter optimization, as illustrated in Fig.\ref{structure}. Firstly, we extract features from the labeled dataset and target dataset through CNNs, respectively. Features from the labeled dataset are defined as $X_l\in \mathbb{R}^{h\times w\times c}$, and features from the target dataset are defined as  $X_t\in \mathbb{R}^{h\times w\times c}$, where $c$ is the number of channels, $h$ and $w$ are the height and width of the feature map, respectively. Then, the feature maps $X_l$  and $X_t$ are projected into high-level graph representation $Y_l\in \mathbb{R}^{N\times d}$ and $Y_t\in \mathbb{R}^{N\times d}$. Here $N= h\times w$ denotes the vertices of the graph, and $d$ denotes the desired feature dimension. The projection can be defined as the function $f(\cdot)$ as:

\begin{equation}
Y_l=f(X_l),
\end{equation}
\begin{equation}
Y_t=f(X_t).
\end{equation}

Subsequently, we leverage a learnable adjacency matrix to encode feature relations by graph reasoning. The graph representations focus on local features, so we carry out graph propagation over the representations $Y_l$ and $Y_t$ to generate the evolved feature $Y_l^n$ and $Y_t^n$ by following graph convolution \cite{Kipf17} as:

\begin{equation}
    Y_l^n=\sigma (A_l^e Y_l^{n-1} W_l^e),
\end{equation}
\begin{equation}
    Y_t^n=\sigma (A_t^e Y_t^{n-1} W_t^e),
\end{equation}
where $W_l^e \in \mathbb{R}^{d\times d}$ and $W_t^e \in \mathbb{R}^{d\times d}$ are trainable weight matrices, $\sigma$ is the ReLU function, $n=1, 2, 3$ represents the first, second, and third graph convolution, respectively. It should be noted that $Y_l^0=Y_l$ and $Y_t^0=Y_t$. The node adjacency weight matrices $A_l^e$ and $A_t^e$ are learnable matrices. They are capable of learning the correlation between different nodes in the graph. We utilize a learnable matrix as the node adjacency weight matrix. In this way, the adjacency matrices $A_l^e$ and $A_t^e$ are randomly initialized, which can be learned during training.

The evolved features $Y_l^n$ and $Y_t^n$ are fused through the inter-graph fusion, resulting in the new target graph feature. To sufficiently propagate global information and produce hierarchical features, graph convolution is implemented several times as shown in Fig. \ref{structure}. In our implementation, three graph convolutions are utilized.

Finally, the evolved features are utilized to boost the image representation. Similar to Eq. 1 and 2, the final graph representation is reprojected to image features. Residual connections \cite{Gong19} are used to further enhance the visual representation with the original feature map to obtain the enhanced feature. The implementation details are shown in Algorithm 1.

\begin{algorithm}[htb]
\caption{The workflow of computing enhanced features by graph-based knowledge supplement network}
\begin{algorithmic}[1]
\Require
Convolution features $X_t$ and $X_l$.
\Ensure
A couple of enhanced features $X_e$.
\State Apply projection to get the high-level graph representation $Y_t$ and $Y_l$
\For{n = 1, 2, 3}
\State Get the evolved feature $Y_l^n$ and $Y_t^n$:
$$Y_l^n=\sigma (A_l^e Y_l^{n-1} W_l^e) \qquad Y_t^n=\sigma (A_t^e Y_t^{n-1} W_t^e) $$
\State Fuse the evolved feature through inter-graph fusion to replace the original $Y_t^n$
$$Y_t^n= \textrm{fusion}(Y_l^n, Y_t^n)$$
\EndFor
\State Adding $Y_t^n$ to the original feature map $X_t$ to form the final enhanced features $X_e$
\end{algorithmic}
\end{algorithm}

\subsection{Inter-Graph Fusion}

To effectively supplement the knowledge extracted from the labeled dataset to the target dataset, a fusion module is essential to distill relevant semantics attentively from one source graph to another target graph. The straightforward solution is to pose them as different branches and combine them directly. However, the underlying contextual information and feature correlations are ignored.

\begin{figure}[ht]
\begin{center}
\includegraphics [width=3.3in]{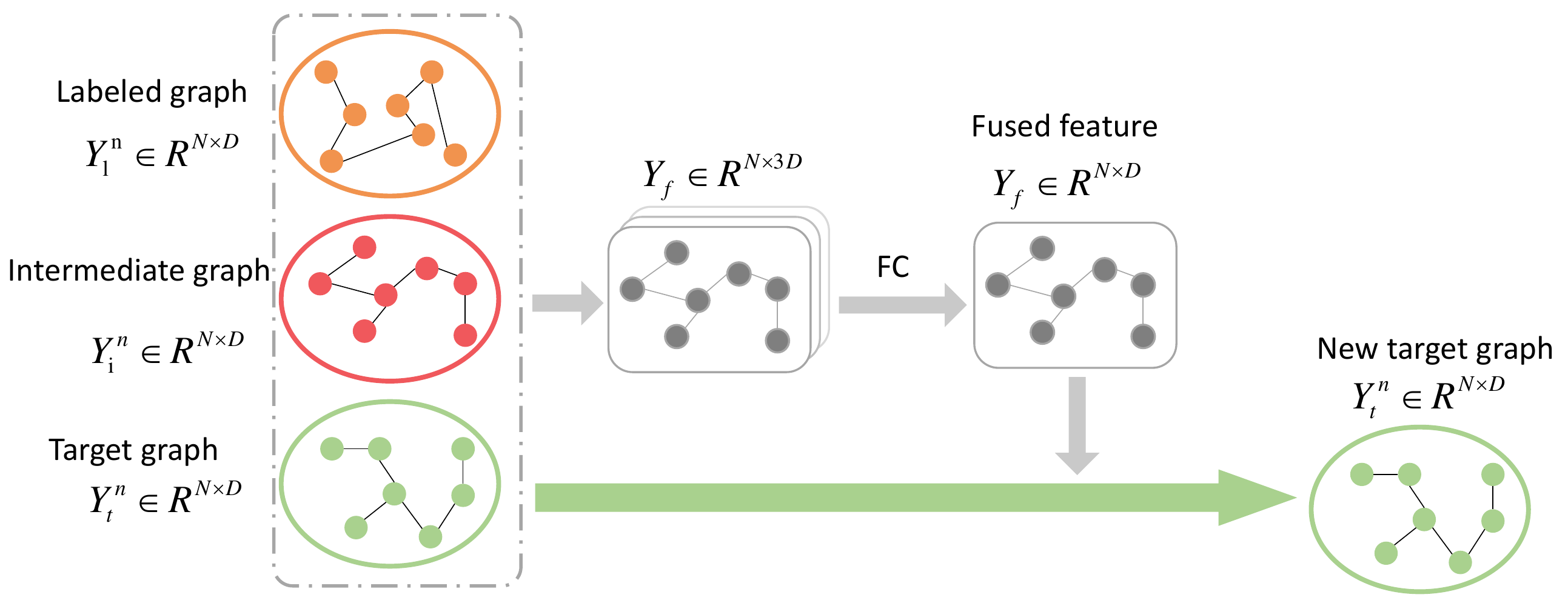}
\caption{Illustration of Inter-graph fusion.}
\label{fig_gdf_module}
\end{center}
\end{figure}

In this paper, we design a graph dependency fusion module to bridge the features of different datasets, as shown in Fig. \ref{fig_gdf_module}. The module can learn proper knowledge dependencies among different datasets. The relationships of vertices from different graphs can be encoded by a similarity matrix as fusion dependency.

Let $G_l = (V_l, E_l)$ denote the labeled graph and $G_t = (V_t, E_t)$ denote the target graph. The graph is represented by a matrix $Y \in \mathbb{R}^{N\times d}$, where $N$ is the number of vertices in the graph, and $d$ is the feature dimension. After graph convolution, the evolved graph features  $Y_t^n$ and $Y_l^n$ can be obtained from the graph representation $Y_t$ and $Y_l$, respectively. Subsequently, graph dependency fusion is used to fuse the evolved graph features which can be formulated as:

\begin{equation}
\textrm{fusion}(Y_l^n, Y_t^n)=Y_t^n+\sigma( \textrm{FC} (Y_t^n, Y_i^n, Y_l^n),
\end{equation}
where FC represents the fully connected layer, and $\sigma$ is the ReLU function. Here $Y_i^n$ is the intermediate graph, which represents a transition from the labeled graph to the target graph. The direct connection between the labeled graph to the target graph may ignore or dilute feature correlations to some extent. Therefore, the intermediate graph is introduced to enhance the feature correlations, which can be calculated as:

\begin{equation}
Y_i^n=A_{tr}Y_l^n W_{i},
\end{equation}
where $W_{i}\in \mathbb{R}^{d \times d}$ is a trainable weight matrix. $A_{tr}=a_{i,j}, i= [1,N_t], j=[1,N_l]$ is a transfer matrix, which is defined according to feature similarity between vertices of $Y_t$ and $Y_l$, where $N_t$ and $N_l$ represents the number of vertices of $Y_t$ and $Y_l$, respectively. The node adjacency weight $a_{i,j}$ can be calculated as:

\begin{equation}
a_{i,j}=\frac{\exp(\cos(v_i,v_j))}{\sum_j \exp(\cos(v_i,v_j))}.
\end{equation}
where $\cos(v_i, v_j)$ is the cosine similarity between $v_i$ and $v_j$.  $v_i$ is the feature of the $i^{th}$ node of corresponding graph, and $v_j$ is the feature of the $j^{th}$ node.

With the well-defined dependency matrix, the labeled graph knowledge and target graph features can be fused and propagated by graph convolution, as expressed in Eq. 3 and 4. Accordingly, the supplementary knowledge and extracted features can be associated and propagated via the inter-graph fusion, which promotes the whole network to generate enhanced features for change detection. After the enhanced features are acquired, they are fed into a classifier consisting of two fully connected layers. The first fully connected layer is used to map features into a low-dimensional feature space, followed by a fully connected layer to map features into changed or unchanged classes, so as to obtain the final change detection results.

\subsection{Embedded Feature Enhancement Model}

\begin{figure*}
\centering
\begin{center}
\includegraphics [width=5.5in]{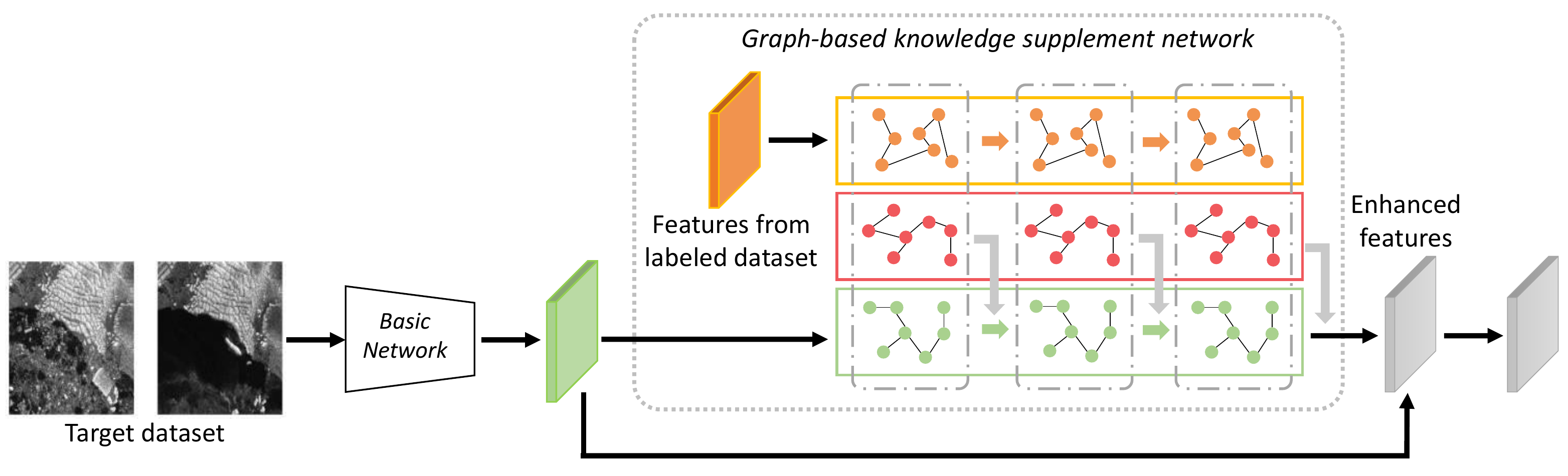}
\caption{Schematic illustration of the proposed method embedded into other networks.}
\label{embedded}
\end{center}
\end{figure*}

As shown in Fig \ref{embedded}, the proposed model can be combined with other CNN-based models. Since it do not change the size of input, the proposed model can be embedded in any convolutional layer. To eliminate the influence of noisy samples, the number of training samples may not meet the needs of parameter optimization when only the features of target dataset are used. With the help of intra-graph reasoning and inter-graph fusion, the proposed GKSNet can alleviate the problem of noisy samples and stabilize the parameter optimization during joint training. When GKSNet is combined with an existing CNN-based network, the extracted features will be enhanced through intra-graph reasoning and inter-graph fusion.

Another merit of the proposed GKSNet is the capability of training on two datasets simultaneously in an end-to-end way. Benefiting from the use of intra-graph reasoning and inter-graph fusion, features from two datasets can be trained simultaneously, rather than fine-tuning on the target dataset after training on the labeled dataset, which is different from other transfer learning-based studies.

\section{Experimental Results and Discussions}

\subsection{Dataset Description and Evaluation Criteria}

The proposed method is validated on four multitemporal SAR datasets. Since the ground truth data is essential for accuracy assessment, the ground truth change maps were manually annotated carefully with expert knowledge. It should be noted that geometric corrections and coregistration have been conducted on these datasets.

\begin{figure}[ht]
\centering
\includegraphics [width=3.3in]{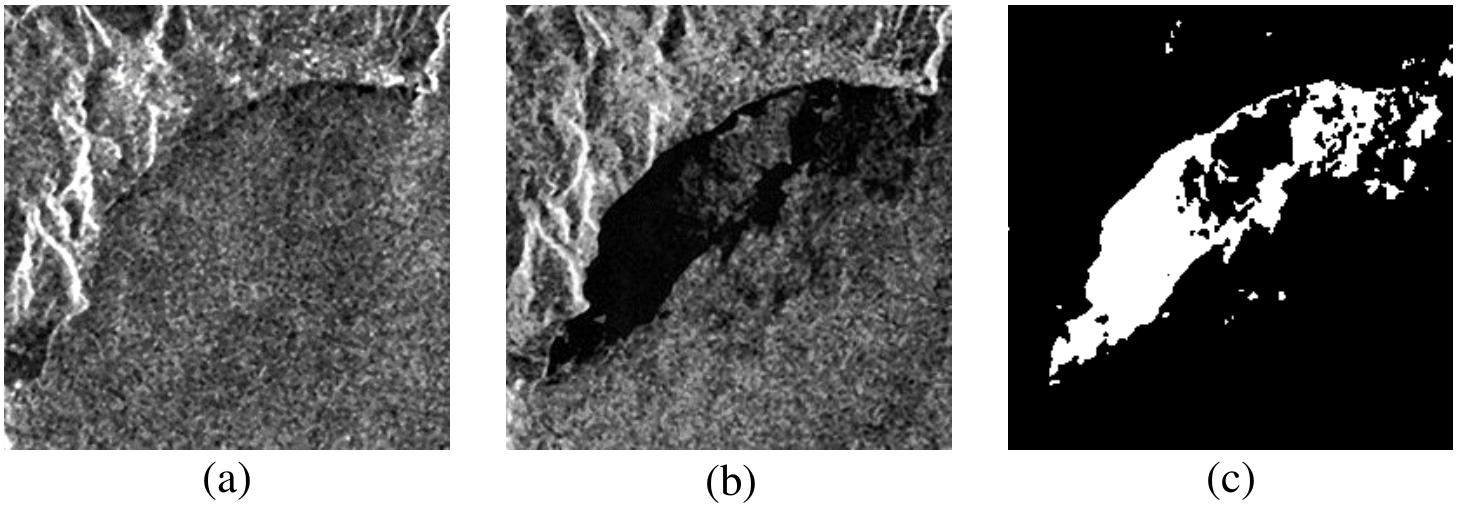}
\caption{Rome dataset. (a) Image captured in April 2003. (b) Image captured in June 2003. (c) Ground truth image.}
\label{Rome}
\end{figure}

The first dataset is the Rome dataset. As illustrated in Fig. \ref{Rome}, the images are captured over an area near Rome, Italy, by the European Remote Sensing (ERS-2) satellite SAR sensor and has the size of 256$\times$256 pixels. The images were collected in April 2003 and June 2003, respectively. The spatial resolution of the dataset is 25 m $\times$ 25 m. The ground-truth change map is annotated by experts with prior knowledge and photo interpretation, as shown in Fig. \ref{Rome}(c).

\begin{figure}[ht]
\centering
\includegraphics [width=3.4in]{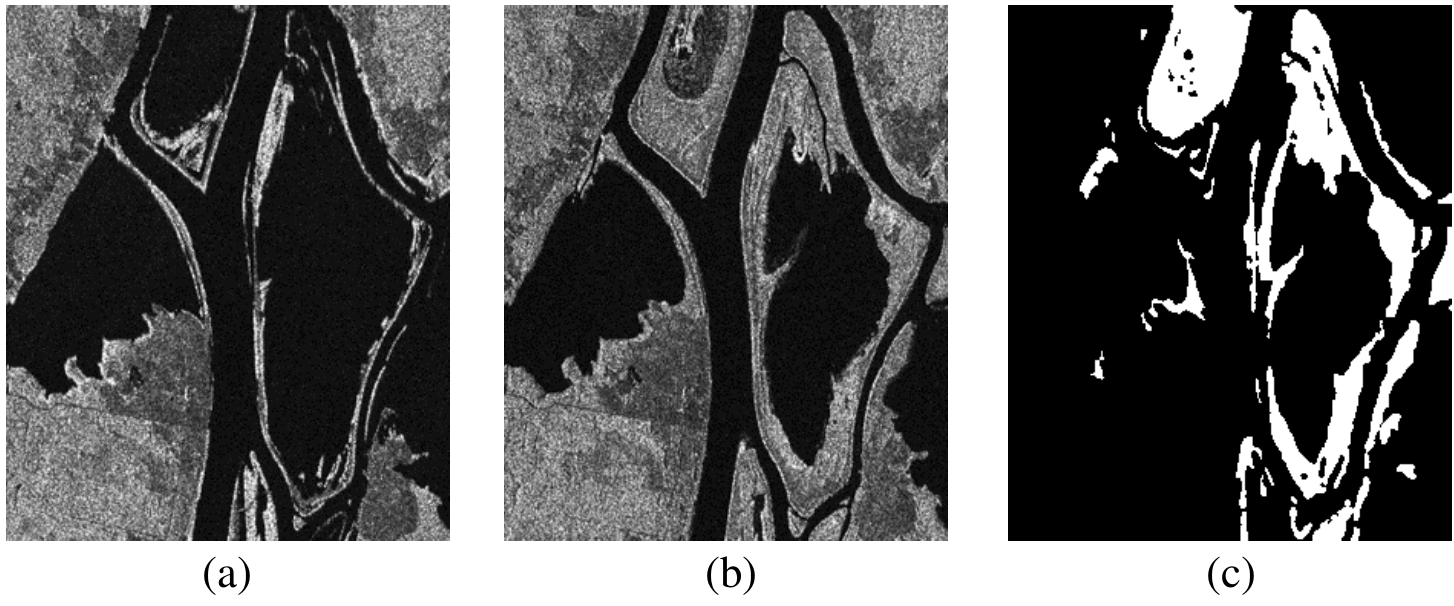}
\caption{Ottawa dataset. (a) Image captured in May 1997. (b) Image captured in August 1997. (c) Ground truth image.}
\label{Ottawa}
\end{figure}

The second dataset is the Ottawa dataset. The images were captured by the Radarsat sensor in May 1997 and August 1997, respectively. As illustrated in Fig. \ref{Ottawa}, the dataset contains images captured over the city of Ottawa. The images were provided by the National Defense Research and Development Canada, and it shows the changed information in areas affected by floods. The available ground truth image is generated by integrating rich knowledge and photo interpretation. The spatial resolution of Ottawa dataset is 10m, and the size is 290$\times$ 350 pixels.

\begin{figure}[ht]
\centering
\includegraphics [width=3.4in]{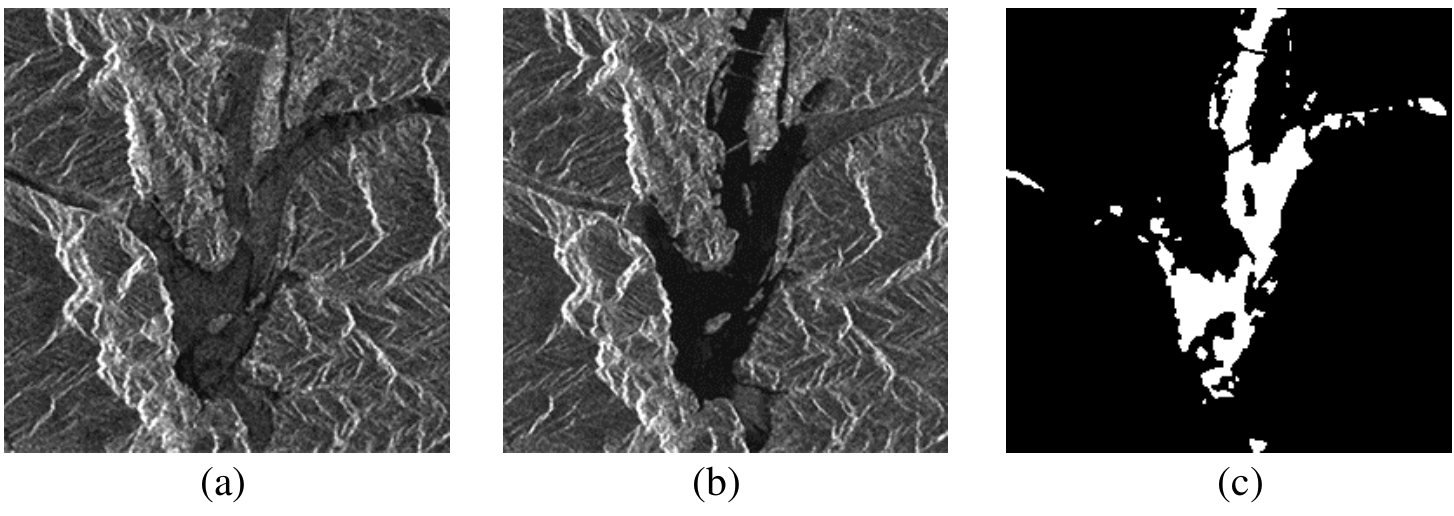}
\caption{Seoul dataset. (a) Image captured in August 2002. (b) Image captured in October 2002. (c) Ground truth image.}
\label{Seoul}
\end{figure}

The third dataset is the Seoul dataset, which is shown in Fig. \ref{Seoul}. Both images were captured around Seoul by the ERS-2 satellite in August 2002 and October 2002, respectively. One typical region of 256$\times$ 256 pixels is chosen to demonstrate the efficacy of the proposed GKSNet. The spatial resolution of Seoul dataset is 25m, which reflects the change of river level before and after precipitation.

\begin{figure}[ht]
\centering
\includegraphics [width=3.4in]{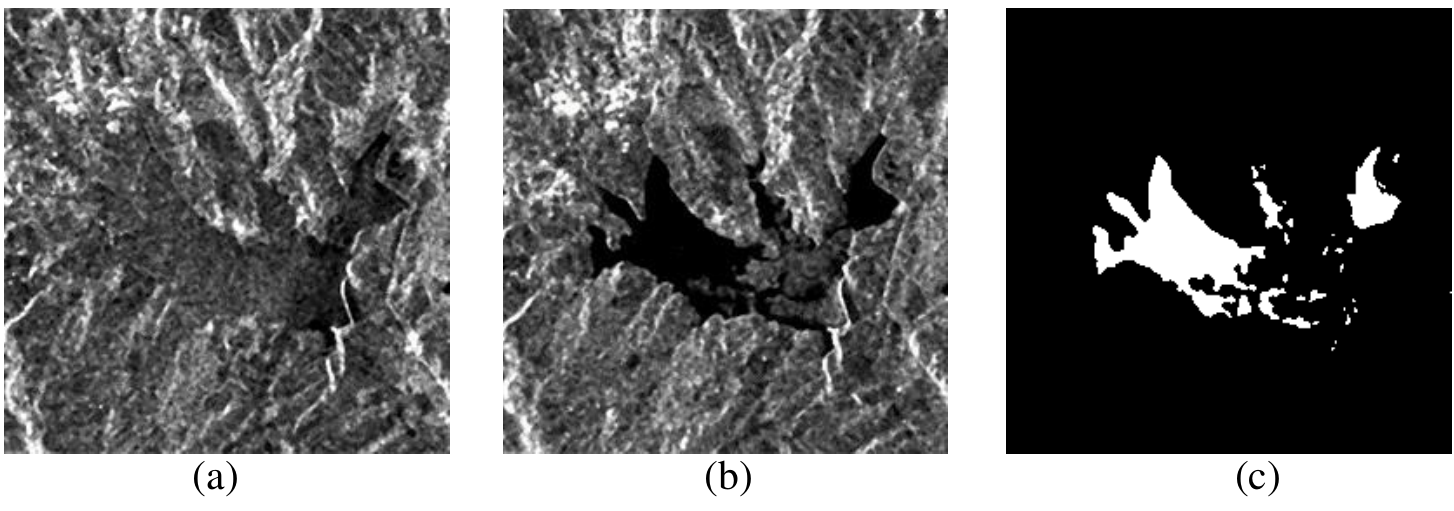}
\caption{Florence dataset. (a) Image captured in July 2004. (b) Image captured in September 2004. (c) Ground truth image.}
\label{Florence}
\end{figure}

The fourth dataset is the Florence dataset (Fig. \ref{Florence}), which was captured over the city of Florence, Italy with 25m spatial resolution. Some parts of the river have changed during the acquisition time. Both images are captured in July 2004 and September 2004, respectively, by the ERS-2 satellite SAR sensor. To better display the change information, an area with the size of 256$\times$ 256 pixels is selected. The ground truth image is shown in Fig. \ref{Florence}(c).

\begin{figure}[ht]
\centering
\includegraphics [width=3.4in]{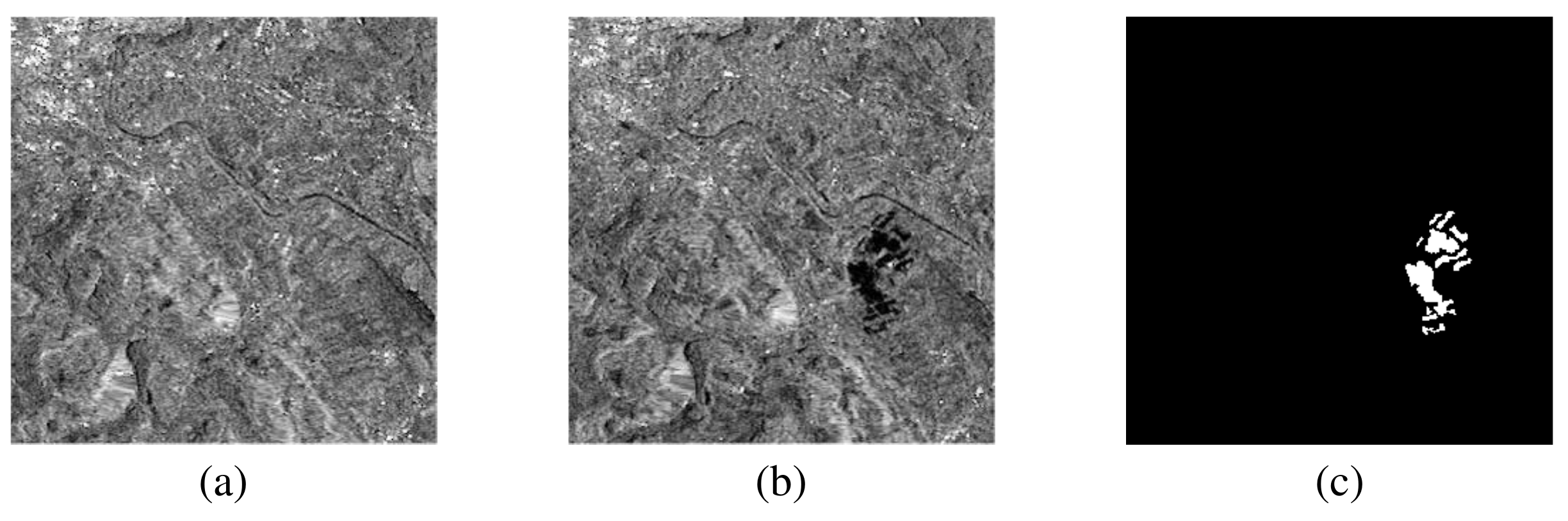}
\caption{Bern dataset. (a) Image captured in April 1999. (b) Image captured in May 1999. (c) Ground truth image.}
\label{Bern}
\end{figure}

The last dataset is the Bern dataset (Fig. \ref{Bern}), which was captured over an area near the city of Bern, Switzerland in April and May 1999, respectively. Two images show the geomorphic changes after the river Aare flooded parts of the cities of Thun and Bern and the airport of Bern entirely. A section (301$\times$ 301 pixels) of two SAR images acquired by the ERS-2 satellite SAR sensor was chosen to verify the effectiveness of various changed detection methods. The ground truth image is shown in Fig. \ref{Bern}(c).

Quantitative evaluation indices including false positives (FP), false negatives (FN), overall errors (OE), percentage correct classification (PCC), Kappa coefficient (KC) and F1 score (F1) are used to evaluate the proposed method. The FP is the number of pixels which are unchanged in the ground truth image but falsely identified as changed in the change detection result. The FN denotes the number of pixels which are changed in the ground truth but falsely identified as unchanged in the change detection result. The TP is the number of pixels which are changed in the ground truth image and truly classified as changed. $N_u$ represents the number of unchanged pixels in the ground truth image, and $N_c$ represents the number of changed pixels in the ground truth image. Then, the OE can be computed by using $\textrm{OE} = \textrm{FP} + \textrm{FN}$.  The PCC can be computed by:
\begin{equation}
\textrm{PCC} = \frac{N_c+N_u-\textrm{OE}}{N_c+N_u}\times 100\%.
\end{equation}

KC can be formulated as:
\begin{equation}
\textrm{KC} = \frac{\textrm{PCC}-\textrm{PRE}}
{1 - \textrm{PRE}} \times 100\%,
\end{equation}
\begin{equation}
\textrm{PRE}=\frac{
(N_c+\textrm{FP}-\textrm{FN})\times N_c +
(N_u+\textrm{FN}-\textrm{FP})\times N_u}{
(N_c+N_u) \times (N_c+N_u)}.
\end{equation}

F1 can be formulated as:
\begin{equation}
\textrm{F1} = \frac{\textrm{2TP}}{\textrm{2TP}+\textrm{FP}+\textrm{FN}},
\end{equation}

All our experiments are implemented on one NVIDIA GeForce 2080Ti GPU. The model is trained for 300 epochs. The first 100 epochs maintained a learning rate of 0.0001, and the learning rate is decayed by a factor of 0.5 every 50 epochs.

\subsection{Number of Graph Convolution}

In the proposed GKSNet, Intar-graph reasoning is a critical component. To sufficiently propagate global information and extract hierarchical features, we need to analyze the number of graph convolutions. The first experiment tests the graph convolution number $n$ in the GKSNet. The performance of change detection is evaluated by taking $n$ = 1, 2, 3, 4, and 5. The corresponding PCC values are employed as the validation criterion.

\begin{figure}[]
\centering
\includegraphics [width=3.2in]{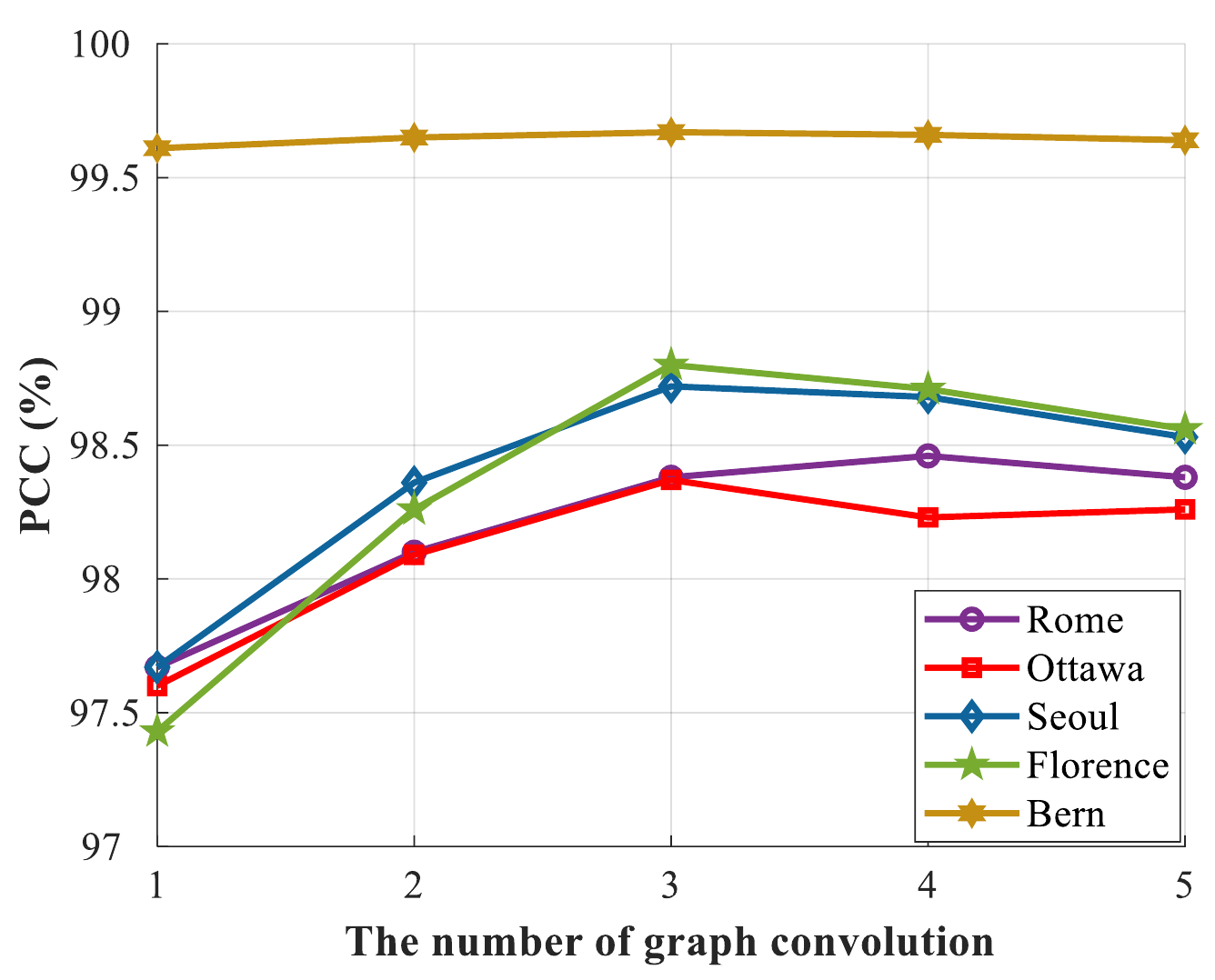}
\caption{Relationship between PCC and the number of graph convolutions.}
\label{graphconvolution}
\end{figure}

Fig. \ref{graphconvolution} shows the quantitative analysis result of the graph convolution number $n$ in the proposed GKSNet. It can be observed that when $n=3$, the best performance is achieved on the Ottawa, Seoul, Florence and Bern datasets. On the Rome dataset, the best results are achieved when $n=4$. With the increase of the graph convolution number, the changed information can be better identified with more global information. However, in most cases, when $n>3$, stacking more layers in the GKSNet leads to the over smoothing of the output features. This is because graph convolution, as a low-pass filter, tends to homogenize the features of different nodes when multiple layers of graph convolution are stacked. In addition, fewer training samples are more likely to cause such problems. Therefore, considering the computational efficiency and accuracy, $n$ is set to 3 in our following experiments.

\begin{figure}[]
\centering
\includegraphics [width=3.0in]{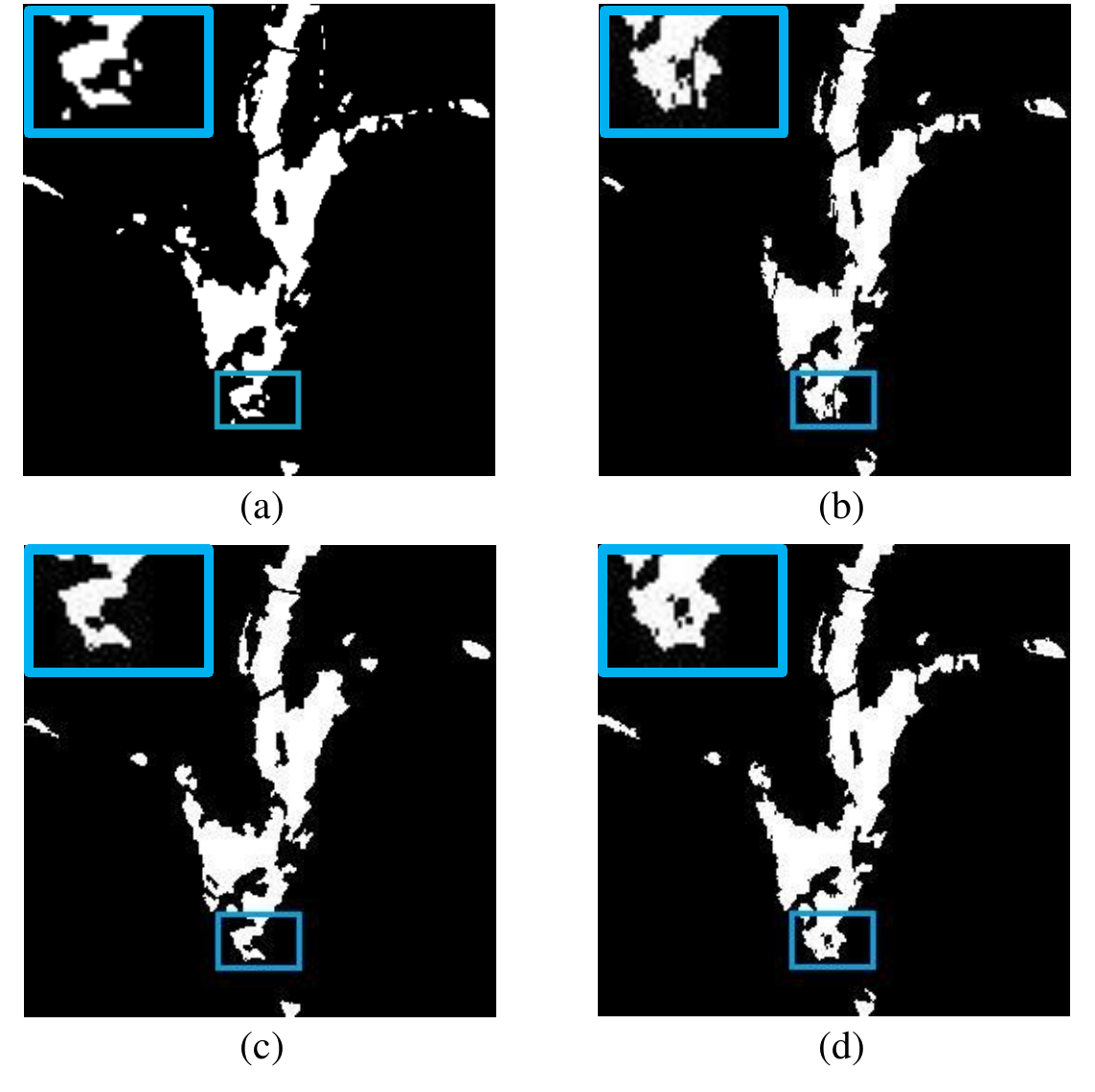}
\caption{Change detection results on Seoul dataset with different $n$ values. (a) Ground-truth image. (b) $n$ = 1. (c) $n$ = 3. (d) $n$ = 5.}
\label{graph}
\end{figure}

Visual comparison of different $n$ values on the Seoul dataset is illustrated in Fig. \ref{graph}. It can be observed that when $n=3$, the generated change map is the most similar to the ground truth. When $n>3$, multiple graph convolution will lead to excessive dissemination of information, which results in the confusion of detailed features. The regions marked in the blue box in Fig. \ref{graph} demonstrate the improvements when $n=3$.

\subsection{Number of Training Samples}

Deep learning-based methods commonly require a large number of samples for parameter optimization. Therefore, the number of training samples is a critical parameter in the proposed model. In this paper, we intend to suppress noisy samples while ensuring robust features. Hence, unlike other methods \cite{Gong16_tnnls}\cite{Gao16_grsl}\cite{Gao19_rs} which generally take about 10\% of the total pixels in the dataset as training samples, the proposed GKSNet requires fewer training samples. Accordingly, the proposed GKSNet effectively reduces the impact of less reliable samples.

\begin{figure}[ht]
\centering
\includegraphics [width=3.2in]{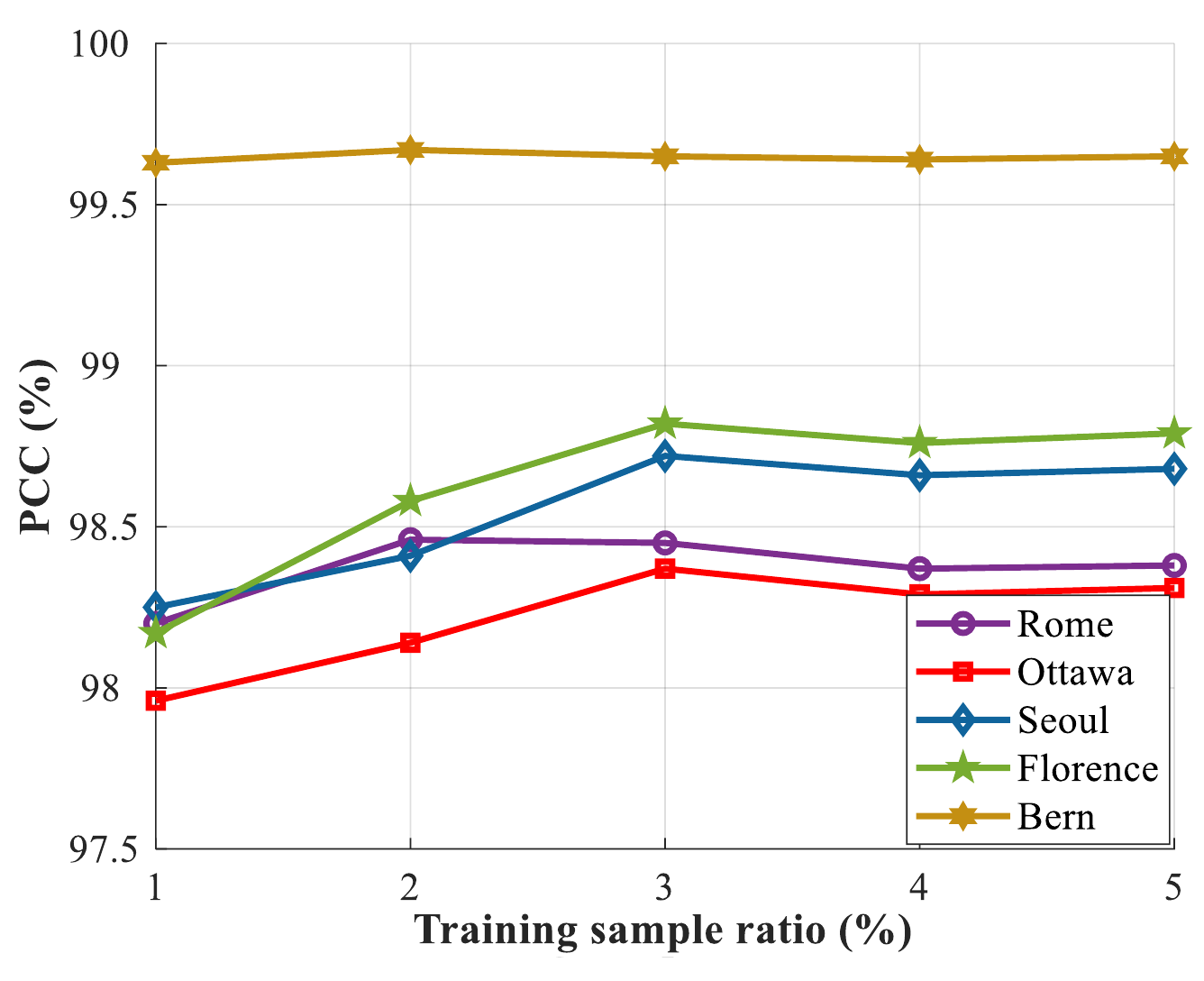}
\caption{Relationship between PCC and the number of training samples.}
\label{trainratio}
\end{figure}

We selected 1\%, 2\%, 3\%, 4\% and 5\% pixels as training samples. Fig. \ref{trainratio} shows the relationship between PCC values and training sample numbers on five datasets. It can be observed that the PCC values can reach a satisfying level when only 3\% training samples are used for most datasets. For the Rome and Bern dataset, the best result is achieved when the training sample number ratio is 2\%. After that, the PCC value tends to be stable when the number of training samples grows. Thus, we select 2\% pixels as the training samples on the Rome and Bern dataset, and 3\% on the other datasets. This ratio can not only obtain good performance, but also help to achieve computational efficiency. As mentioned before, the training samples of some existing methods are generally 10\%, which is 3 to 5 times greater than the proposed GKSNet. It is evident that the proposed GKSNet is capable of exploiting the common knowledge and does not require a large number of training samples.

\subsection{Analysis of the Patch Size}

The size of the patch is an important parameter that controls the spatial contextual information contained in the input data. When the size is small, the spatial information contained in the data is insufficient, which leads to the lack of discrimination of the samples;  When the size is large, additional interference information will inevitably be introduced, which affects the final result. Therefore, to verify the effect of different sizes of patches on the final change detection result, relevant experiments are carried out in this subsection to select the optimal patch size. Let $r$ denote the size of patches taken from the original image for feature extraction. We evaluate the change detection performance by taking $r$ = 3, 5, 7, 9, 11, and 13.

\begin{figure}[ht]
\centering
\includegraphics [width=3.2in]{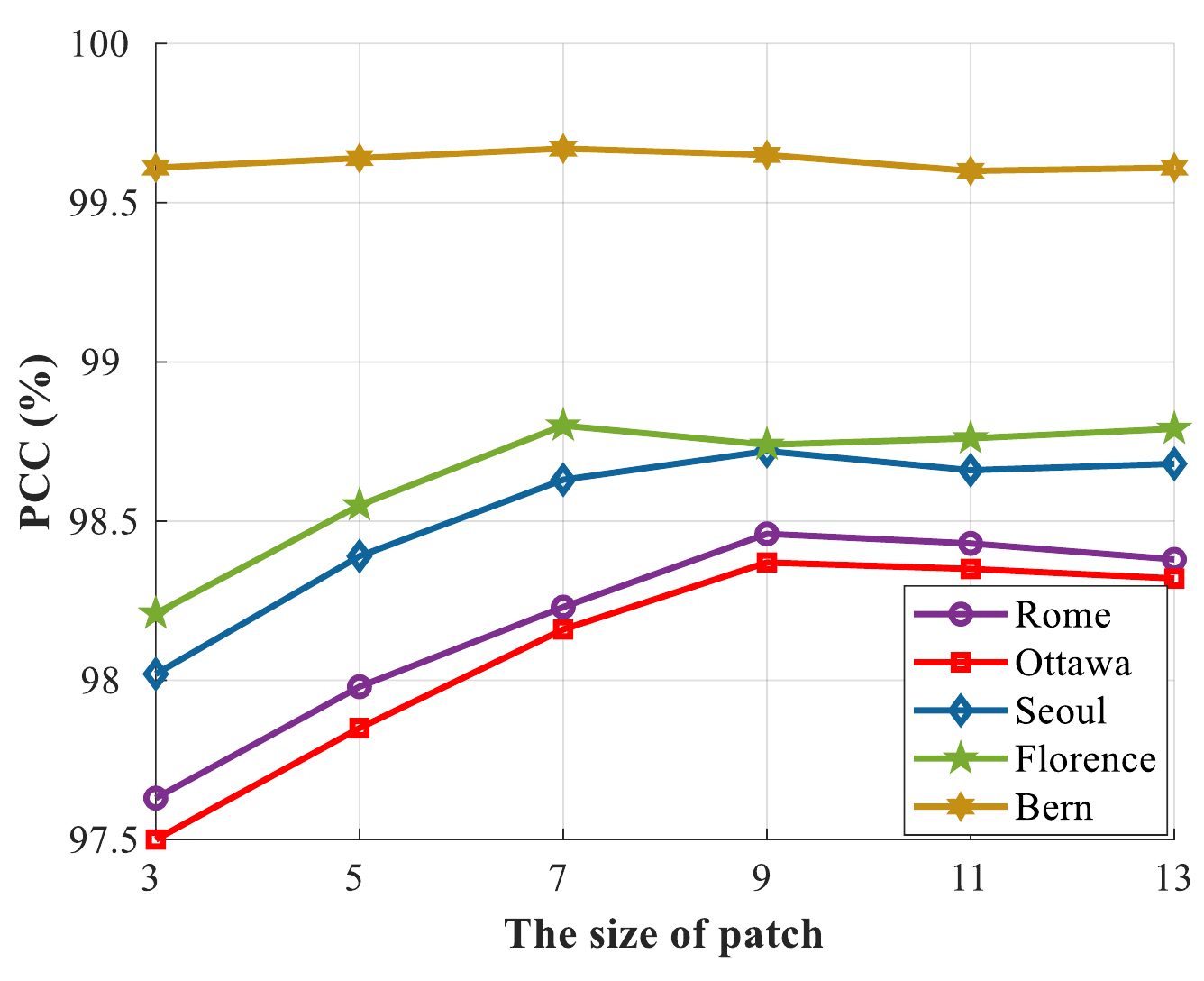}
\caption{Relationship between PCC and the size of patch.}
\label{patchsize}
\end{figure}

As illustrated in Fig.\ref{patchsize}, we can see that the PCC value is not satisfying when $r<=7$, because small patches may not contain enough contextual information for discrimination. On the Florence and Bern datasets, the GKSNet achieves the best result when $r=7$. While on the Rome, Ottawa, and Seoul datasets, the proposed method achieves the best performance when $r=9$. It indicates that different image scenes require different receptive fields for feature extraction. Therefore, in our implementation, we set $r=7$ on the Florence and Bern datasets and $r=9$ on the Rome, Ottawa, and Seoul datasets.

\subsection{Combinations of Labeled and Target Datasets}

The proposed GKSNet aims to extract discriminative information from the labeled dataset, and then this information is employed to supplement knowledge to the target dataset. However, due to different data distribution, different combinations of labeled and target datasets will generate different results. Table \ref{table_combine} illustrates the PCC values by employing different combinations.

\begin{table}[htb]
\caption{The change detection results of different combination of the labeled dataset and target dataset.}
\renewcommand\arraystretch{1.50}
\centering
\setlength{\tabcolsep}{1.5mm}{
\begin{tabular}{c|l|ccccc}
\hline\hline
\multicolumn{2}{l|}{\multirow{2}{*}{}}                                                & \multicolumn{4}{c}{Labeled} \\ \cline{3-7}
\multicolumn{2}{l|}{}
& ~Rome~  & ~Ottawa~  & ~Seoul~  & ~Florence~  & ~Bern~  \\ \hline
\multirow{4}{*}{Target}
& Rome     & ~---~ & 98.28 & \textbf{98.46} & 97.93 & 97.84 \\
& Ottawa   & 98.15 & ~---~ & 98.13 & \textbf{98.37} & 98.21 \\
& Seoul    & \textbf{98.72} & 97.36 & ~---~ & 98.44 & 97.45\\
& Florence & 98.61 & \textbf{98.80} & 98.44 & ~---~ & 98.78 \\
& Bern     & 99.64 & 99.66 &99.64 & \textbf{99.67} & ~---~ \\
\hline\hline
\end{tabular}
}
\label{table_combine}
\end{table}

From Table \ref{table_combine}, it can be observed that different combination of the labeled datasets and target datasets can produce various change detection results. Since image feature distribution varies among different datasets, datasets with similar distributions commonly perform better in knowledge supplements. The Ottawa, Florence and Bern datasets reflect the surrounding areas of the city, while the Seoul and Rome datasets account for more natural landforms, which provides a basis for their complementary features. In addition, it can be seen that GSKNet still produces competitive results with fewer training samples, even if it is not the optimal combination, which also proves that the proposed method can effectively transfer knowledge among datasets.

\subsection{Experimental Results and Discussion}

To validate the performance of the proposed GKSNet, we compare it with several state-of-the-art methods, including PCAKM \cite{Celik09}, NR-ELM \cite{Gao16_jars}, GaborPCANet \cite{Gao16_grsl}, LR-CNN \cite{Liu19}, MLFN \cite{gao19_grsl}, DBN \cite{Gong16_tnnls} and DCNet \cite{Gao19_rs}. For PCAKM, the contextual information is analyzed by principal component analysis, and the extracted features are clustered by $k$-means algorithm. NR-ELM utilizes the neighborhood-based ratio operator to obtain reliable training samples. Then, ELM is employed to train a model by using these samples. GaborPCANet is a simplified deep learning model which is comprised of several PCA layers and binary hashing layers. LR-CNN is formed by imposing a spatial constraint on the output layer of CNN. MLFN proposed a transferred multilevel fusion network, which trained on a large dataset to transfer deep knowledge from the data set to the limited training data. In DBN, a deep belief network is utilized for SAR image change detection task. DCNet establishes a very deep cascade network to exploit discriminative features and introduce a fusion mechanism to combine the output of different hierarchical layers to further alleviate the exploding gradient problem. For fair comparison, compared methods are implemented by using the default parameters. Both visual and quantitative analyses are made to reflect the results of various methods more intuitively and effectively.

\subsubsection{Results on the Rome Dataset}

\begin{figure}[ht]
\centering
\includegraphics [width=3.4in]{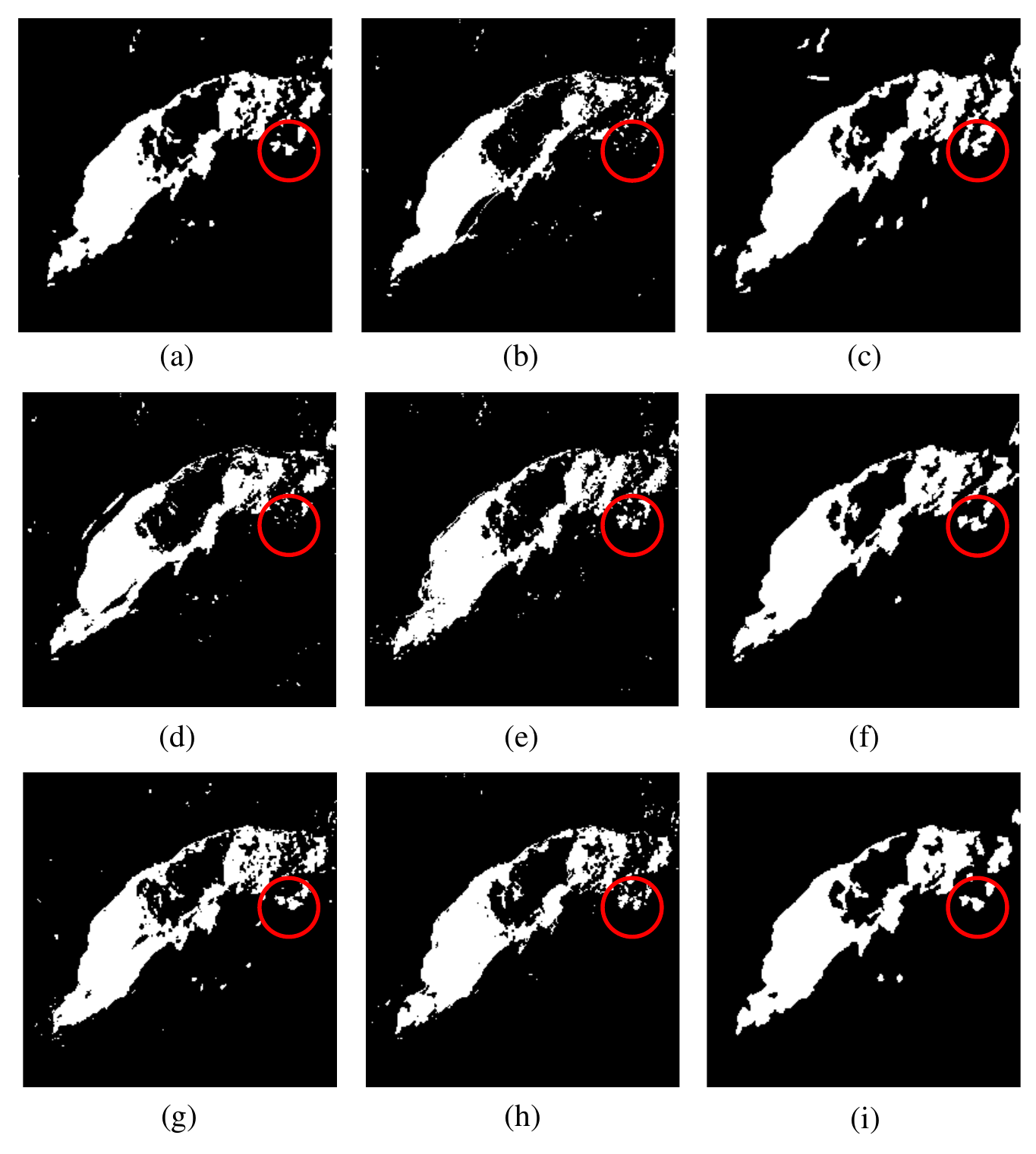}
\caption{Visualized results of different change detection methods on the Rome dataset.(a) Ground truth image. (b) Result by PCAKM. (c) Result by NR-ELM. (d) Result by GaborPCANet. (e) Result by LR-CNN. (f) Result by MLFN. (g) Result by DBN. (h) Result by DCNet. (i) Result by the proposed GKSNet.}
\label{fig_Rome}
\end{figure}

\begin{table}[htbp]
\centering
\begin{center}
\caption{change detection results of different methods on the Rome dataset}
\renewcommand\arraystretch{1.50}
\setlength{\tabcolsep}{1.5mm}{
\begin{tabular}{l|cccccc}
\hline\hline
Methods                        & FP   & FN   & OE   & PCC (\%) & KC(\%) & F1(\%)\\ \hline
PCAKM \cite{Celik09}           & 649  & 1864 & 2513 & 96.17    & 82.93  & 85.12\\
NR-ELM \cite{Gao16_jars}       & 1726 & \textbf{101}  & 1827 & 97.21    & 89.12  & 90.74\\
GaborPCANet \cite{Gao16_grsl}  & 785  & 1348 & 2133 & 96.75    & 85.96 & 87.84 \\
LR-CNN \cite{Liu19}            & 582  & 1057 & 1639 & 97.50    & 89.26  & 90.70\\
MLFN \cite{gao19_grsl}         & 952  & 242  & 1194 & 98.18    & 92.59  & 93.65\\
DBN \cite{Gong16_tnnls}        & 682  & 758  & 1440 & 97.80    & 90.74  & 92.01\\
DCNet \cite{Gao19_rs}          & \textbf{339}  & 888  & 1227 & 98.12    & 91.93  & 93.01\\
GKSNet                & 706  & 304  & \textbf{1010} & \textbf{98.46}    & \textbf{93.64} & \textbf{94.54}\\
\hline \hline
\end{tabular}
}
\label{table_Rome}
\end{center}
\end{table}

From Fig. \ref{fig_Rome}, we can see that the dataset contains complex spatial structures, especially there are some unchanged pixels within the changed area. Besides, the Rome dataset is also seriously interfered by speckle noise. Thus, it is a challenging task to identify changed pixels accurately in this dataset. From Table \ref{table_Rome}, we can see that in addition to NR-ELM and the proposed GKSNet, other methods have high FN values, which means many changed pixels are missed. This phenomenon can be verified from the area marked in the red circle. Meanwhile, the result of NR-ELM suffers from high FP value, which indicates some unchanged areas are falsely detected as changed ones as shown in Fig.\ref{fig_Rome} (c). From visual inspection, we can observe some noisy regions exiting in the results of PCAKM, GaborPCANet, LR-CNN and DBN, which affect the final results. Among all methods, the proposed GKSNet generates the best PCC, KC and F1 values and has the smallest OE value, which is resulted by supplementary knowledge and can be regarded as a suppression of noise in pseudo-label. Moreover, fusing with discriminative features from other datasets, the proposed GKSNet can decrease the influence of speckle noise to some extent. The experimental results show that the proposed GKSNet can make full use of the transferred knowledge to alleviate the impact of noisy samples, and effectively implement change detection on the Rome dataset.

\subsubsection{Results on the Ottawa Dataset}

\begin{figure}[ht]
\centering
\includegraphics [width=3.4in]{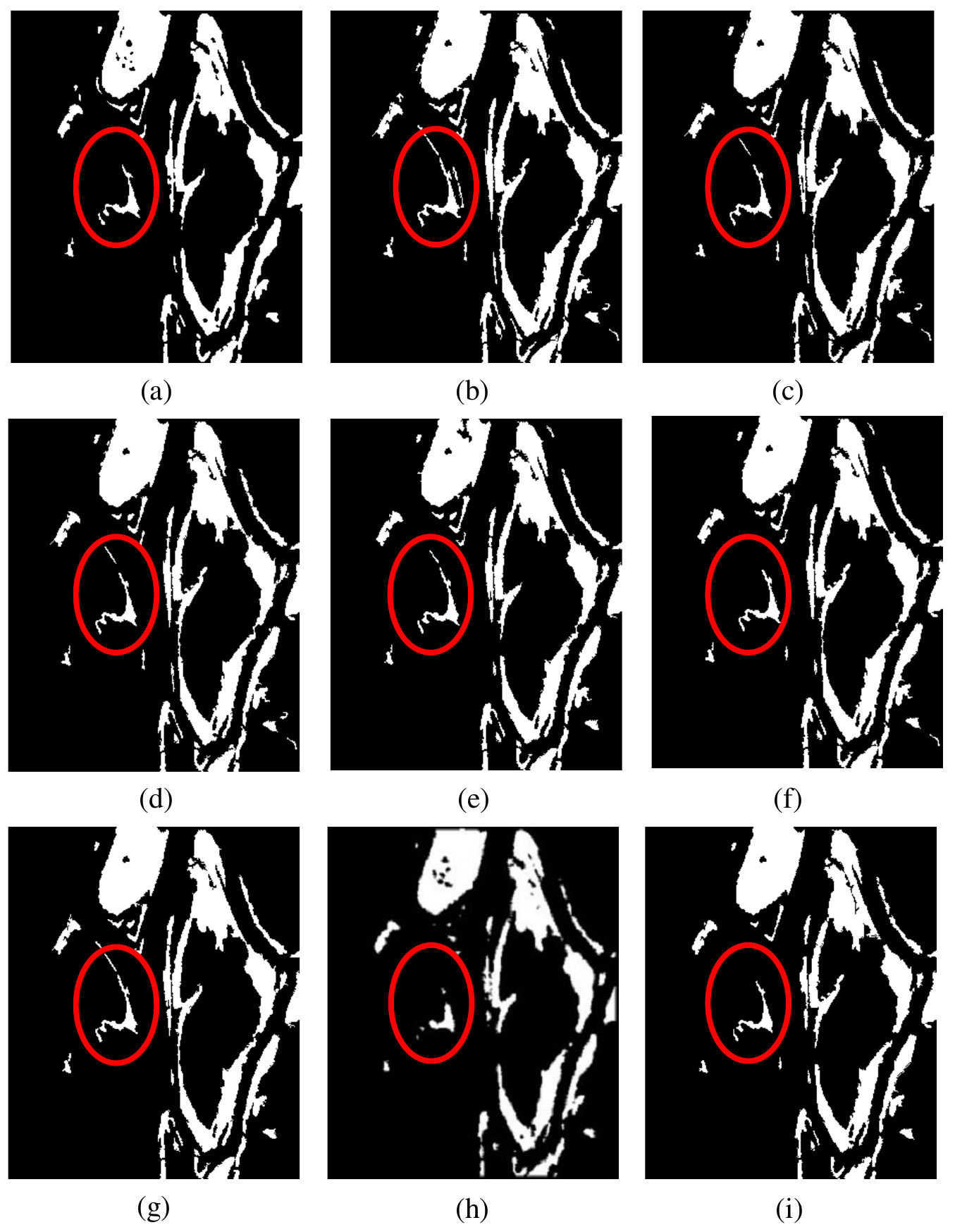}
\caption{Visualized results of different change detection methods on the Ottawa dataset.(a) Ground truth image. (b) Result by PCAKM. (c) Result by NR-ELM. (d) Result by GaborPCANet. (e) Result by LR-CNN. (f) Result by MLFN. (g) Result by DBN. (h) Result by DCNet. (i) Result by the proposed GKSNet.}
\label{fig_Ottawa}
\end{figure}

\begin{table}[htbp]
\centering
\begin{center}
\caption{change detection results of different methods on the Ottawa dataset}
\renewcommand\arraystretch{1.50}
\setlength{\tabcolsep}{1.5mm}{
\begin{tabular}{c|cccccc}
\hline\hline
Methods                         & FP   & FN   & OE   & PCC (\%) & KC(\%) & F1(\%)\\ \hline
PCAKM \cite{Celik09}          & 2191   & \textbf{320}  & 2511 & 97.53  & 91.13  &92.60\\
NR-ELM \cite{Gao16_jars}      & 1333   & 679  & 2012 & 98.02    & 92.68  &93.86\\
GaborPCANet \cite{Gao16_grsl} & 1512   & 534  & 2046 & 97.98    & 92.61 &93.81 \\
LR-CNN \cite{Liu19}           & 1218   & 725  & 1943 & 98.09    & 92.90 &94.04 \\
MLFN \cite{gao19_grsl}        & 1509   & 445  & 1954 & 98.07    & 92.96 &93.97 \\
DBN \cite{Gong16_tnnls}       & 1347   & 561  & 1908 & 98.12    & 93.08 &94.20 \\
DCNet \cite{Gao19_rs}         & \textbf{762}  & 1068 & 1830 & 98.20    & 93.18 &94.12 \\
GKSNet                        & 825  & 829  & \textbf{1654} & \textbf{98.37}   & \textbf{93.88}  &\textbf{94.80}\\
\hline \hline
\end{tabular}
}
\label{table_Ottawa}
\end{center}
\end{table}

Fig.\ref{fig_Ottawa} illustrates the change detection result by different methods on the Ottawa dataset. The corresponding evaluation metrics are listed in Table \ref{table_Ottawa}, where it can be observed that for PCAKM, NR-ELM, GaborPCANet, LR-CNN, MLFN and DBN suffer from high FP values, resulting in some noisy areas as marked in red circles. Moreover, deep learning-based methods (MLFN, DBN, DCNet and the proposed GKSNet) achieve better performance than shallow models. Based on visual comparisons, it is obvious that the proposed GKSNet provides more similar results to the ground truth in the marked area. Specifically, in the marked area, PCAKM, NR-ELM, GaborPCANet, LR-CNN and DBN generate extra changed region. DCNet ignores some detailed information, and many changed pixels are missed. However, the changed detection result by the proposed GKSNet is the most similar to the ground truth change map. Table \ref{table_Ottawa} shows that GKSNet yields the best PCC value of 98.37\%, which has increased by 0.17\% at least compared with other methods. It is evident that the proposed GKSNet can exploit the structure similarity and underlying common knowledge from different datasets. The comparisons demonstrate the superior performance of the proposed method on the Ottawa dataset.

\subsubsection{Results on the Seoul Dataset}

\begin{figure}[ht]
\centering
\includegraphics [width=3.4in]{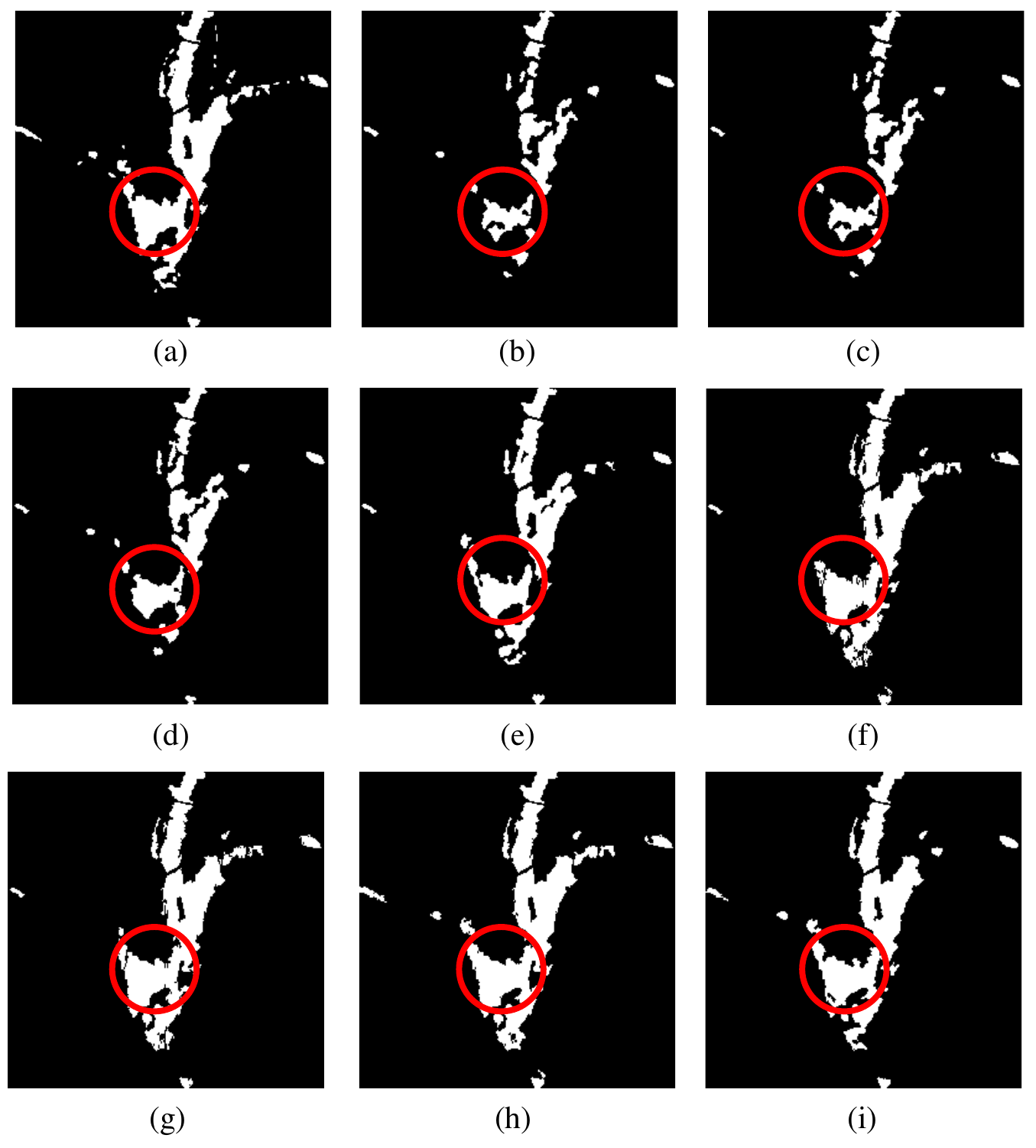}
\caption{Visualized results of different change detection methods on the Seoul dataset.(a) Ground truth image. (b) Result by PCAKM. (c) Result by NR-ELM. (d) Result by GaborPCANet. (e) Result by LR-CNN. (f) Result by MLFN. (g) Result by DBN. (h) Result by DCNet. (i) Result by the proposed GKSNet.}
\label{fig_Seoul}
\end{figure}

\begin{table}[htbp]
\centering
\begin{center}
\caption{change detection results of different methods on the Seoul dataset}
\renewcommand\arraystretch{1.50}
\setlength{\tabcolsep}{1.5mm}{
\begin{tabular}{c|cccccc}
\hline\hline
Methods                       & FP   & FN   & OE   & PCC (\%) & KC(\%) & F1(\%)\\ \hline
PCAKM \cite{Celik09}          & 892  & 1832 & 2724 & 95.84    & 73.10  & 70.45\\
NR-ELM \cite{Gao16_jars}      & 864  & 1908 & 2772 & 95.77    & 72.39  & 69.93\\
GaborPCANet \cite{Gao16_grsl} & 1043 & 1009 & 2052 & 96.87    & 81.21  & 79.39\\
LR-CNN \cite{Liu19}           & 654  & 525  & 1179 & 98.20    & 89.28  & 89.34\\
MLFN \cite{gao19_grsl}        & 611  & 535  & 1146 & 98.25    & 89.54  & 89.70\\
DBN \cite{Gong16_tnnls}       & 618  & 476  & 1094 & 98.33    & 90.06  & 90.98\\
DCNet \cite{Gao19_rs}         & 768  & \textbf{194}  & 962  & 98.53    & 91.63  & 92.34\\
GKSNet                        & \textbf{408} & 428  & \textbf{836}  & \textbf{98.72} & \textbf{92.31}  &\textbf{93.02}\\
\hline \hline
\end{tabular}
}
\label{table_Seoul}
\end{center}
\end{table}

Fig. \ref{fig_Seoul} presents the change detection results on the Seoul dataset. The corresponding quantitative metrics are listed in Table \ref{table_Seoul}. The results of PCAKM, NR-ELM and GaborPCANet miss many changed regions, thus these methods suffer from very high FN values. For LR-CNN, MLFN, DBN and DCNet, the values of FP are relatively high since some unchanged regions are generally divided into changed regions, which can be partly attributed to noisy samples. In contrast, GKSNet effectively avoids the influence of noisy samples and improves the experimental performance by introducing additional knowledge from other data. Moreover, it can be seen that deep learning-based methods (MLFN, DBN, DCNet and GKSNet) perform better than classical shallow models. Compared with DBN, the KC value of the proposed GKSNet has increased by 2.25\%. Compared with DCNet, the KC value of the proposed GKSNet has increased by 0.68\%. This demonstrates that the proposed GKSNet is suitable for Seoul dataset. From the marked regions in Fig. \ref{fig_Seoul}, we can observe that the changed area of the results of  PCAKM, NR-ELM, GaborPCANet and LR-CNN is clearly reduced, which is consistent with the evaluation metrics in Table \ref{table_Seoul}. Besides, the results of MLFN, DBN and DCNet generate many changed pixels, which results in contour inconsistent with the ground-truth map. Among all methods, the result of GKSNet is much closer to the ground-truth map. This further shows that the GKSNet can well integrate the features of different datasets to enhance the ability of feature representation for satisfying change detection results.

\subsubsection{Results on the Florence Dataset}

\begin{figure}[ht]
\centering
\includegraphics [width=3.4in]{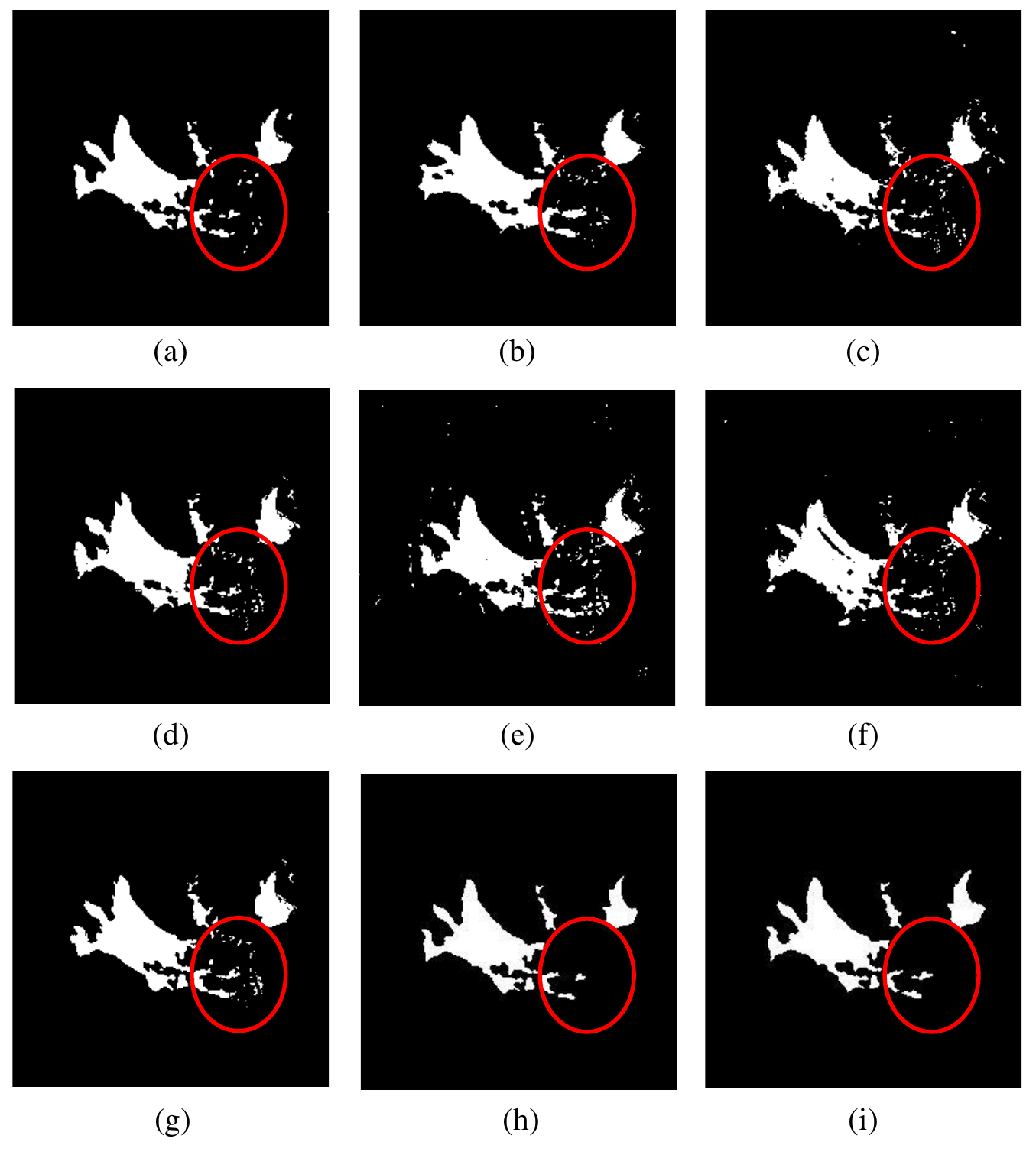}
\caption{Visualized results of different change detection methods on the Florence dataset.(a) Ground truth image. (b) Result by PCAKM. (c) Result by NR-ELM. (d) Result by GaborPCANet. (e) Result by LR-CNN. (f) Result by MLFN. (g) Result by DBN. (h) Result by DCNet. (i) Result by the proposed GKSNet.}
\label{fig_Florence}
\end{figure}

\begin{table}[htbp]
\centering
\begin{center}
\caption{change detection results of different methods on the Florence dataset}
\renewcommand\arraystretch{1.50}
\setlength{\tabcolsep}{1.5mm}{
\begin{tabular}{c|cccccc}
\hline\hline
Methods                        & FP   & FN   & OE   & PCC (\%) & KC(\%) & F1(\%)\\ \hline
PCAKM \cite{Celik09}           & 1426 & 535  & 1961 & 97.01    & 79.13  & 82.20\\
NR-ELM \cite{Gao16_jars}       & 1273 & 520  & 1793 & 97.26    & 80.67  & 82.88\\
GaborPCANet \cite{Gao16_grsl}  & 1055 & 564  & 1619 & 97.53    & 82.11  & 84.86\\
LR-CNN \cite{Liu19}            & 776  & 714  & 1490 & 97.73    & 82.84  & 85.76\\
MLFN \cite{gao19_grsl}         & 702  & 397  & 1099 & 98.32    & 87.64  & 88.54\\
DBN \cite{Gong16_tnnls}        & 796  & \textbf{371}  & 1167 & 98.22    & 87.03  &88.53\\
DCNet \cite{Gao19_rs}          & 454  & 602  & 1056 & 98.39    & 87.58  & 88.67\\
GKSNet                         & \textbf{209} & 576 & \textbf{785} & \textbf{98.80} & \textbf{90.56} &\textbf{91.95} \\
\hline \hline
\end{tabular}
}
\label{table_Florence}
\end{center}
\end{table}

Fig.\ref{fig_Florence} shows the change detection results on the Florence dataset, while Table \ref{table_Florence} lists the evaluation metrics. From Table\ref{table_Florence}, we can observe that PCAKM, NR-ELM, GaborPCANet and DBN suffer from high FP values, which causes many noisy points in the result map. For LR-CNN and DCNet, some changed pixels are missed, and the FN values are relatively high. Moreover, compared with PCAKM, NR-ELM, GaborPCANet and LR-CNN, the performance of deep learning-based methods(MLFN, DBN, DCNet and GKSNet) has been greatly improved. The KC values of MLFN, DBN and DCNet have increased 4.80\%, 4.19\% and 4.64\%, respectively, compared with the shallow models. The proposed GKSNet has improved with a value of 7.72\%. For more intuitive comparisons, we have marked a region in Fig.\ref{fig_Florence}. It can be noticed that there are many false alarms in the results of PCAKM, NR-ELM, GaborPCANet and DBN, which is consistent with the higher FP values in Table \ref{table_Florence}. For LR-CNN and DCNet, they missed many changed pixels. Among these methods, the proposed GKSNet has the best PCC, KC and F1 values and generates the best change map, which is very similar to the reference ground truth map. The comparison shows that the proposed GKSNet is powerful in discriminative feature extraction and is effective on the Florence dataset.

\subsubsection{Results on the Bern Dataset}

\begin{figure}[ht]
\centering
\includegraphics [width=3.4in]{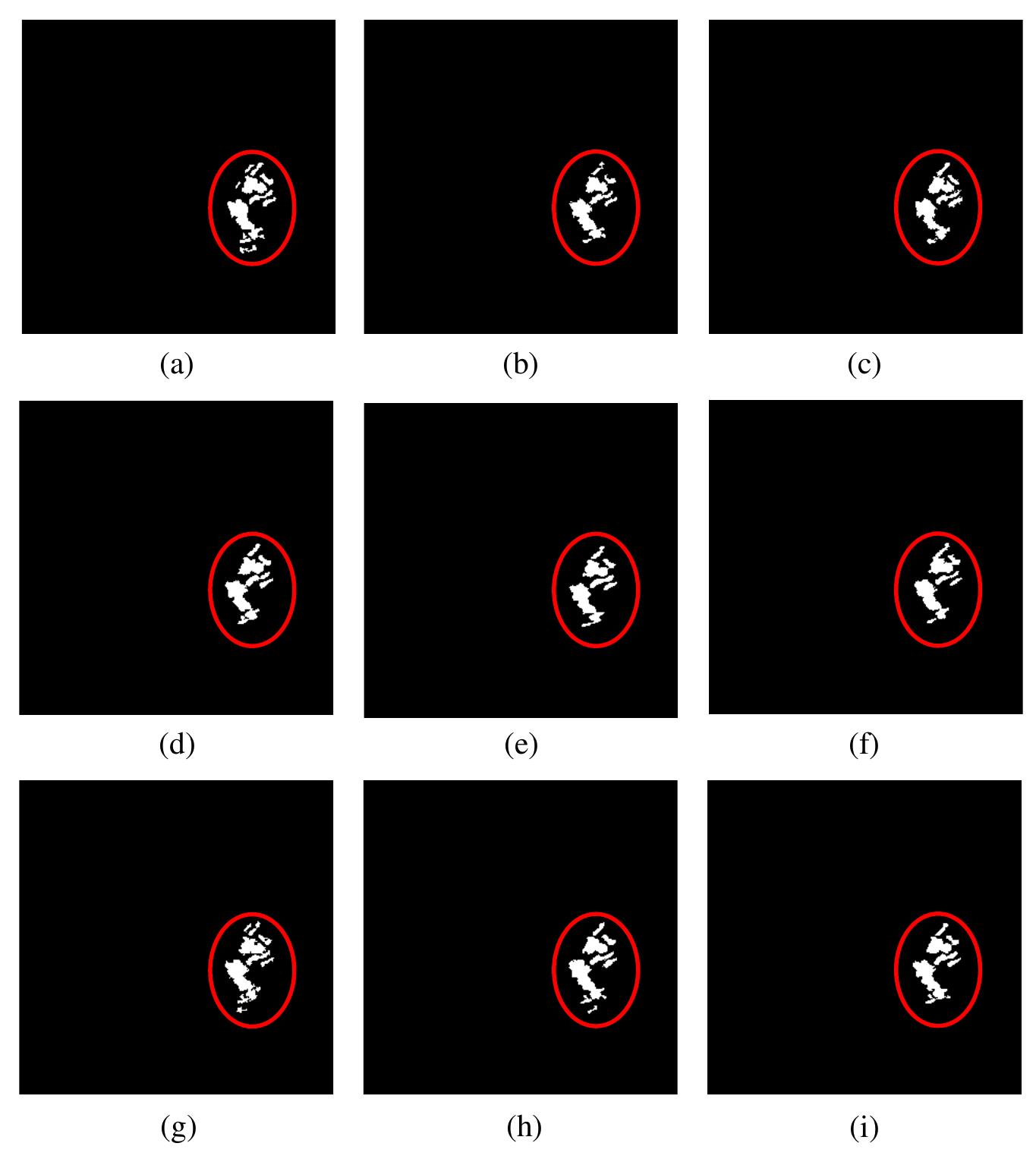}
\caption{Visualized results of different change detection methods on the Bern dataset.(a) Ground truth image. (b) Result by PCAKM. (c) Result by NR-ELM. (d) Result by GaborPCANet. (e) Result by LR-CNN. (f) Result by MLFN. (g) Result by DBN. (h) Result by DCNet. (i) Result by the proposed GKSNet.}
\label{fig_Bern}
\end{figure}

\begin{table}[htbp]
\centering
\begin{center}
\caption{change detection results of different methods on the Bern dataset}
\renewcommand\arraystretch{1.50}
\setlength{\tabcolsep}{1.5mm}{
\begin{tabular}{c|cccccc}
\hline\hline
Methods                        & FP   & FN   & OE   & PCC (\%) & KC(\%) & F1(\%)\\ \hline
PCAKM \cite{Celik09}           & \textbf{95}  & 258  & 353 & 99.61    & 83.36  & 83.56\\
NR-ELM \cite{Gao16_jars}       & 104 & 228  & 332 & 99.63    & 84.63  & 84.81\\
GaborPCANet \cite{Gao16_grsl}  & 131 & 193  & 324 & 99.64    & 85.41  & 85.59\\
LR-CNN \cite{Liu19}            & 97  & 216  & 313 & 99.65    & 85.54  & 85.71\\
MLFN \cite{gao19_grsl}         & 97  & 225  & 322 & 99.64    & 85.06  & 85.24\\
DBN \cite{Gong16_tnnls}        & 145 & \textbf{159}  & 304 & 99.66    & \textbf{86.59}  & \textbf{86.76}\\
DCNet \cite{Gao19_rs}          & 146 & 161  & 307 & 99.66    & 86.45  & 86.62\\
GKSNet                         & 111 & 190 & \textbf{301} & \textbf{99.67} & 86.34 & 86.51 \\
\hline \hline
\end{tabular}
}
\label{table_Bern}
\end{center}
\end{table}

The visualization of change detection results on the Bern dataset is shown in Fig.\ref{fig_Bern}, and the quantitative evaluation metrics are listed in Table \ref{table_Bern}. From Table \ref{table_Bern}, we can see that GKSNet has the smallest OE value, which means that it can produce result with the least misclassification. Moreover, the satisfactory results achieved by the proposed method on this open source dataset demonstrate the effectiveness of GKSNet. The visualisation results also confirm this conclusion. The generated change detection map is similar to the ground truth map, which proves that the proposed method can effectively deal with the noise in pseudo-label, and transfer meaningful knowledge from other datasets.

Based on the experiments on these real SAR datasets, the proposed GKSNet offers better performance over classical models. Besides, by combining with the supplementary knowledge, the proposed GKSNet yields superior performance over other deep learning-based methods in most cases with less training samples. Moreover, the inter-graph fusion exploits the feature similarity and underlying common knowledge among different datasets, which further improves the change detection performance.

\subsection{Ablation Studies}

\begin{table}[]
\centering
\begin{center}
\caption{Details of the Basic Network}
\renewcommand\arraystretch{1.50}
\setlength{\tabcolsep}{1.3mm}{
\begin{tabular}{ccccc}
\hline\hline
Layer & ~~ Input ~~ & Type  & Kernel size \\ \hline
1 & $r\times r\times3$ & Conv / BN / ReLU & $3\times3$ \\
2 & $r\times r\times64$ & Conv / BN / ReLU & $3\times3$ \\
3 & $r\times r\times64$ & Conv / BN / ReLU & $3\times3$ \\
4 & $r\times r\times64$ & Conv / BN / ReLU & $1\times1$ \\
5 & $r\times r\times128$ & Conv / BN / ReLU & $1\times1$ \\
6 & $r\times r\times128$ & Sum &  \\
7 & $r\times r\times128$ & ReLU &   \\
\hline\hline
\end{tabular}
}
\label{backbone}
\end{center}
\end{table}

The proposed GKSNet is employed to obtain enhanced features conditioned on a simple but complete network, called basic network. The basic network is illustrated in Table. \ref{backbone} which is used to extract the initial features. To further discuss and validate the effectiveness of the two components of GKSNet, we have made ablation studies on the five SAR datasets. We designed three variants: 1) \textbf{Basic Network} denotes the backbone, which does not absorb the knowledge supplement provided in additional datasets, but trains and tests on one dataset. 2) \textbf{w/o Inter-Graph Fusion} refers to the GKSNet using intra-graph convolution to generate the evolved features but combining the graph convolutional features directly, instead of using inter-graph fusion. 3) \textbf{Gaussian Kernel} refers to using Gaussian kernel to measure the similarity between data, instead of consine similarity. 4) \textbf{GKSNet} denotes the complete model.

\begin{table}[htbp]
\centering
\begin{center}
\renewcommand\arraystretch{1.80}
\caption{Ablation Studies on Four Datasets}
\begin{tabular}{c|ccccc}
\hline\hline
\multirow{2}{*}{Method} & \multicolumn{5}{c}{PCC (\%) on different datasets} \\ \cline{2-6}
& Rome & Ottawa & Seoul & Florence & Bern \\ \hline
Basic Network & 97.65 & 97.59 & 95.81 & 96.96 & 99.59 \\
w/o Inter-Graph Fusion  & 98.29 & 98.02 & 97.32 & 98.37 & 99.64\\
Gaussian Kernel  & 98.42 & 98.10 & 98.59 & 98.76 & 99.66\\
GKSNet & \textbf{98.46} & \textbf{98.21} & \textbf{98.72} & \textbf{98.80} & \textbf{99.67}\\
\hline\hline
\end{tabular}
\label{table_ablation}
\end{center}
\end{table}

\begin{figure}[ht]
\centering
\includegraphics [width=3.4in]{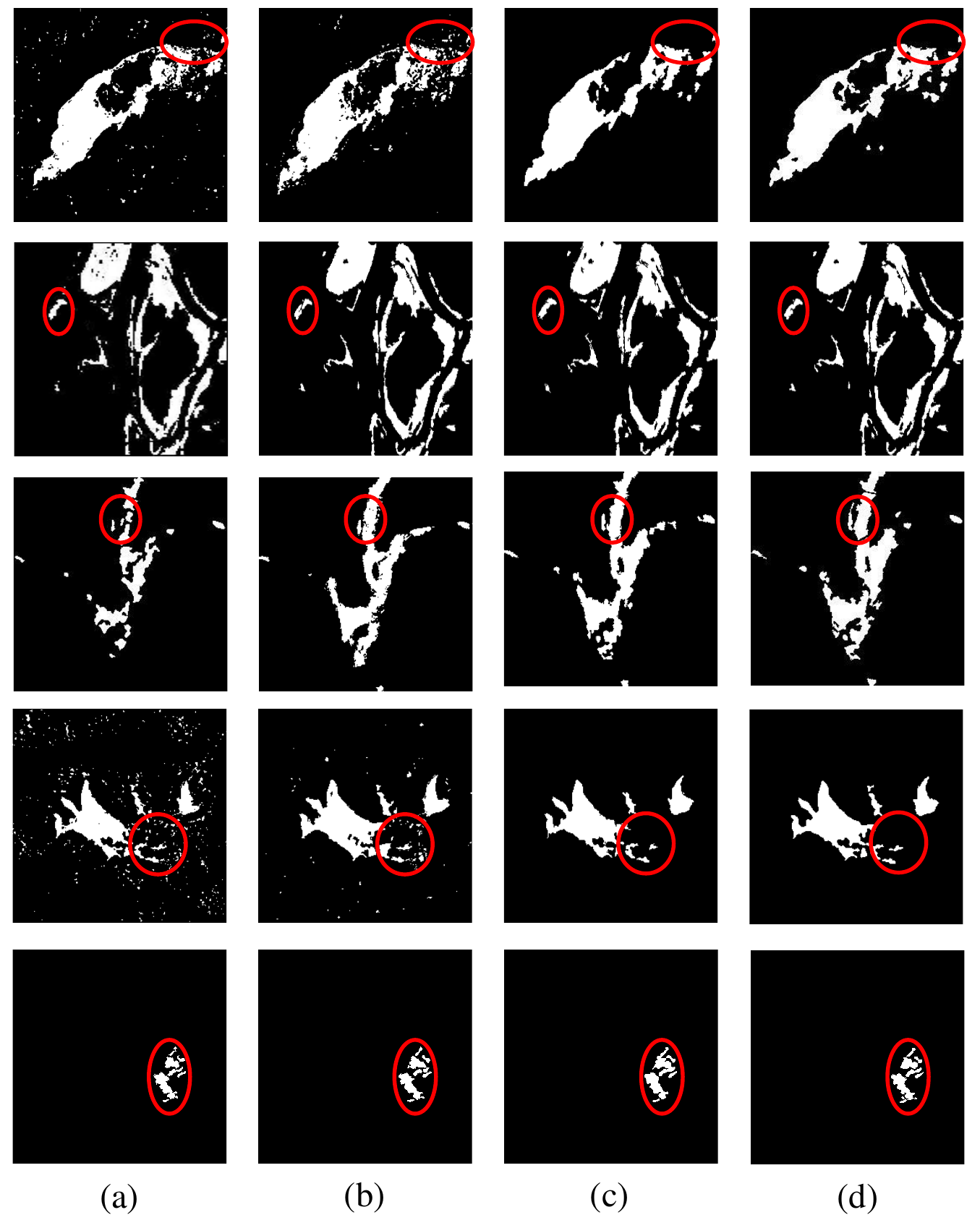}
\caption{Visualized change detection results of variants of our model. (a) Result by Basic Network. (b) Result by w/o Inter-Graph Fusion. (c) Result by Gaussian Kernel. (d) Result by GKSNet.}
\label{ablation}
\end{figure}

As reported in Table \ref{table_ablation}, by combining the supplementary knowledge distilled from other datasets, the result acquires approximately 0.5\% improvements compared with the basic network. The first row shows that without knowledge supplement, a small number of training samples are insufficient to meet the needs of parameter optimization. Furthermore, the proposed GKSNet provides inter-graph fusion mechanism to combine the features from different datasets. By employing the inter-graph fusion, the PCC values increased at least 0.18\%. It is evident that inter-graph fusion promotes change detection performance. It is reasonable that direct feature fusion from different datasets cannot fully explore their correlations, so we introduce the inter-graph fusion which can learn proper feature dependency and knowledge integration among different datasets. Besides, in order to find a more appropriate dependency for inter-graph fusion, we use Gaussian kernel instead of cosine similarity to explore the impact of different data similarity on the experimental results. As can be seen from the Table \ref{table_ablation}, Gaussian kernel can also achieve relatively satisfying performance, but the cosine similarity performs better. Finally, by comparing the result of basic network with Table \ref{table_combine}, it can be observed that even if the data distribution of the labeled dataset is not very similar to the target dataset, it still improve the change detection performance. 

To show the changes brought by GKSNet more intuitively, the results of basic network, w/o Inter-Graph Fusion, Gaussian Kernel and GKSNet are illustrated in Fig. \ref{ablation}. Some obvious differences are marked with red circles. In the case of the Rome dataset, we can see that the result of the basic network contains many noisy regions, since the features are not enough for parameter optimization. By contrast, w/o inter-graph fusion enriches the training features by introducing additional supplementary knowledge, which can better determine the change information of some complex areas, and thus obtain better results. When Gaussian kernel is introduced, it also achieves a good performance. Moreover, the proposed GKSNet not only adds additional knowledge, but combines supplementary knowledge with extracted features. It improves the feature representation and obtains better change detection results.

\begin{figure}[ht]
\centering
\includegraphics [width=3.5in]{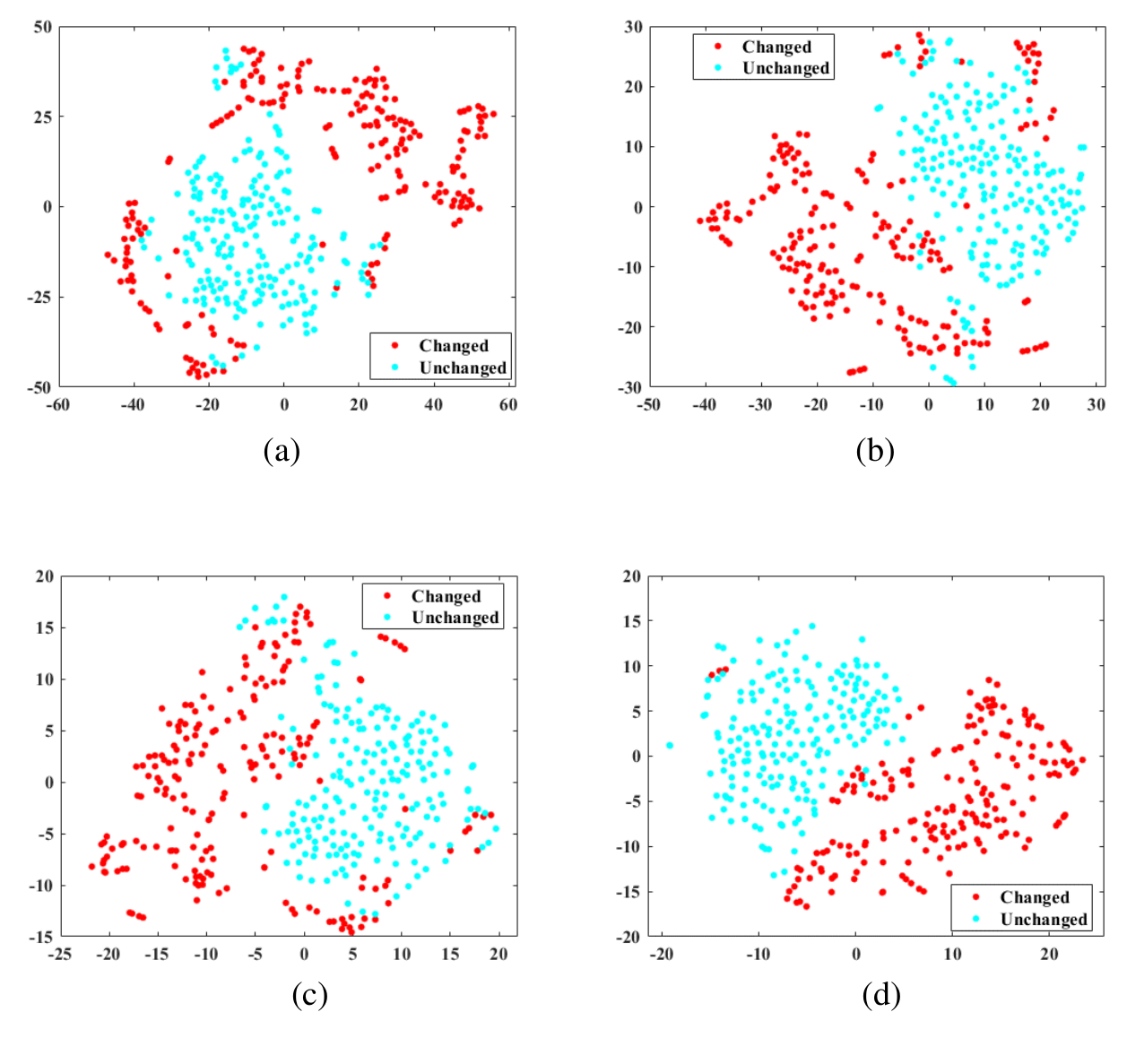}
\caption{Visualization of the feature representations on the Ottawa dataset. (a) Features generated by Basic Network. (b) Features generated by w/o Inter-Graph Fusion. (c) Features generated by Gaussian Kernel. (d) Features generated by GKSNet.}
\label{feature_visual}
\end{figure}

Fig. \ref{feature_visual} visualizes the features generated by four variants. The samples of the same class distribute more closely after intra-graph reasoning and inter-graph fusion. It indicates that the proposed GKSNet achieves the best performance.

\subsection{Runtime Comparisons}

\begin{table}
\centering
\renewcommand\arraystretch{1.80}
\caption{Runtime Comparisons of Different Change Detection Methods}
\begin{tabular}{l|ccccc}
\hline\hline
\multirow{2}{*}{Method} & \multicolumn{5}{c}{Runtime (in seconds)} \\ \cline{2-6}
& Rome & Ottawa & Seoul & Florence & Bern \\ \hline
PCAKM \cite{Celik09} & 2.31 & 2.42 & 2.17 & 2.51 & 2.09\\
NR-ELM \cite{Gao16_jars} & 22.53 & 22.67 & 22.40 & 22.70 & 22.17\\
GaborPCANet \cite{Gao16_grsl} & 435.20 & 447.68 & 432.66 & 469.91 & 428.75 \\
LR-CNN \cite{Liu19} & 281.77 & 293.54 & 272.62 & 298.02 & 266.94\\
MLFN \cite{gao19_grsl} & 187.96 & 196.34 & 174.51 & 205.63 & 173.37\\
DBN \cite{Gong16_tnnls} & 472.34 & 488.95 & 465.90 & 492.01 &451.20 \\
DCNet \cite{Gao19_rs} & 501.02 & 523.64 & 497.76 & 538.47 & 487.16 \\
GKSNet & 147.85 & 148.60 & 136.92 & 157.24 & 133.99\\
\hline\hline
\end{tabular}
\label{timecost}
\end{table}

Time consumption is one of the important factors that restricts the application of deep learning-based methods in change detection. Table \ref{timecost} shows the runtime of the proposed GKSNet with other methods. We can see that traditional methods cost less time because of their relative simple models. For DBN and DCNet, due to the fact that they have complex models and require many training samples for parameter optimization, the time consumption is greater than other methods. However, compared with other deep learning-based methods, the proposed GKSNet is superior in computational time.  The reason is that  GKSNet only takes 20 to 30 percent of the training samples compared with other methods, which greatly reduces the training time. It should be noted that notwithstanding the establishment and propagation of graph network commonly takes about 20s, the proposed GKSNet exhibits high efficiency.

\section{Conclusions}

In this paper, we improve the SAR image change detection by alleviating the effect of noisy samples and utilizing the common knowledge hidden in other datasets. To this end, we proposed a graph-based knowledge supplement network, termed GKSNet. On the one hand, image features from a labeled dataset are projected into a graph. After message propagation via graph convolution, the obtained features are employed as additional knowledge for the target dataset. On the other hand, a graph transfer module is proposed to distill related contextual information attentively from the labeled dataset to the target dataset as supplementary knowledge. The promising experimental results on five real SAR datasets verify the effectiveness of the proposed GKSNet in enhancing feature extraction. In future, attempts will be made to enhance the interpretability of dependencies between mutlitemporal images, and we will to try to reduce the computational burden caused by the graph convolution.

\begin{IEEEbiography}[{\includegraphics[width=1in,height=1.25in,clip,keepaspectratio]{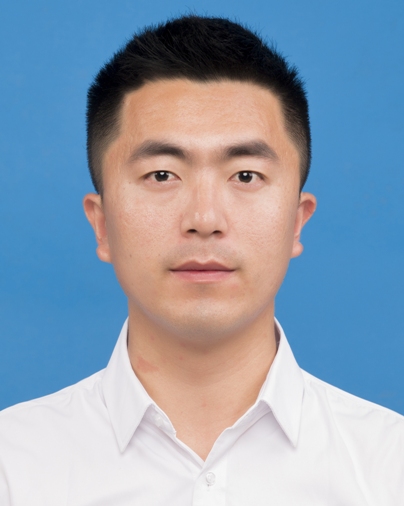}}]{Junjie Wang}
received the B.Sc. degree in computer science from Ocean University of China, Qingdao, China, in 2018. He is currently pursuing the M.Sc. degree in computer science and applied remote sensing with the School of Information Science and Technology, Ocean University of China, Qingdao, China.

His current research interests include computer vision and remote sensing image processing.

\end{IEEEbiography}

\begin{IEEEbiography}[{\includegraphics[width=1in,height=1.25in,clip,keepaspectratio]{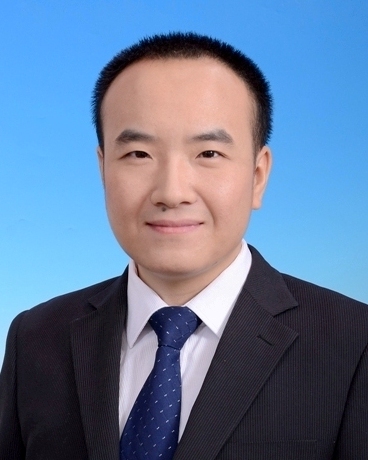}}]{Feng Gao} (Member, IEEE)
received the B.Sc degree in software engineering from Chongqing University, Chongqing, China, in 2008, and the Ph.D. degree in computer science and technology from Beihang University, Beijing, China, in 2015.

He is currently an Associate Professor with the School of Information Science and Engineering, Ocean University of China. His research interests include remote sensing image analysis, pattern recognition and machine learning.

\end{IEEEbiography}

\begin{IEEEbiography}[{\includegraphics[width=1in,height=1.25in,clip,keepaspectratio]{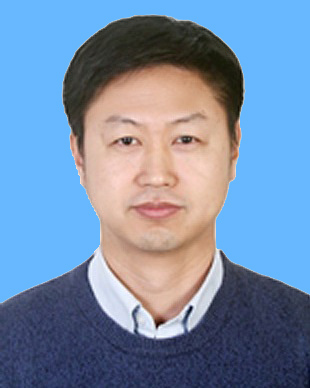}}]{Junyu Dong}
 (Member, IEEE) received the B.Sc. and M.Sc. degrees from the Department of Applied Mathematics, Ocean University of China, Qingdao, China, in 1993 and 1999, respectively, and the Ph.D. degree in image processing from the Department of Computer Science, Heriot-Watt University, Edinburgh, United Kingdom, in 2003.

He is currently a Professor and Dean with the School of Computer Science and Technology, Ocean University of China. His research interests include visual information analysis and understanding, machine learning and underwater image processing.
\end{IEEEbiography}

\begin{IEEEbiography}[{\includegraphics[width=1in,height=1.25in,clip,keepaspectratio]{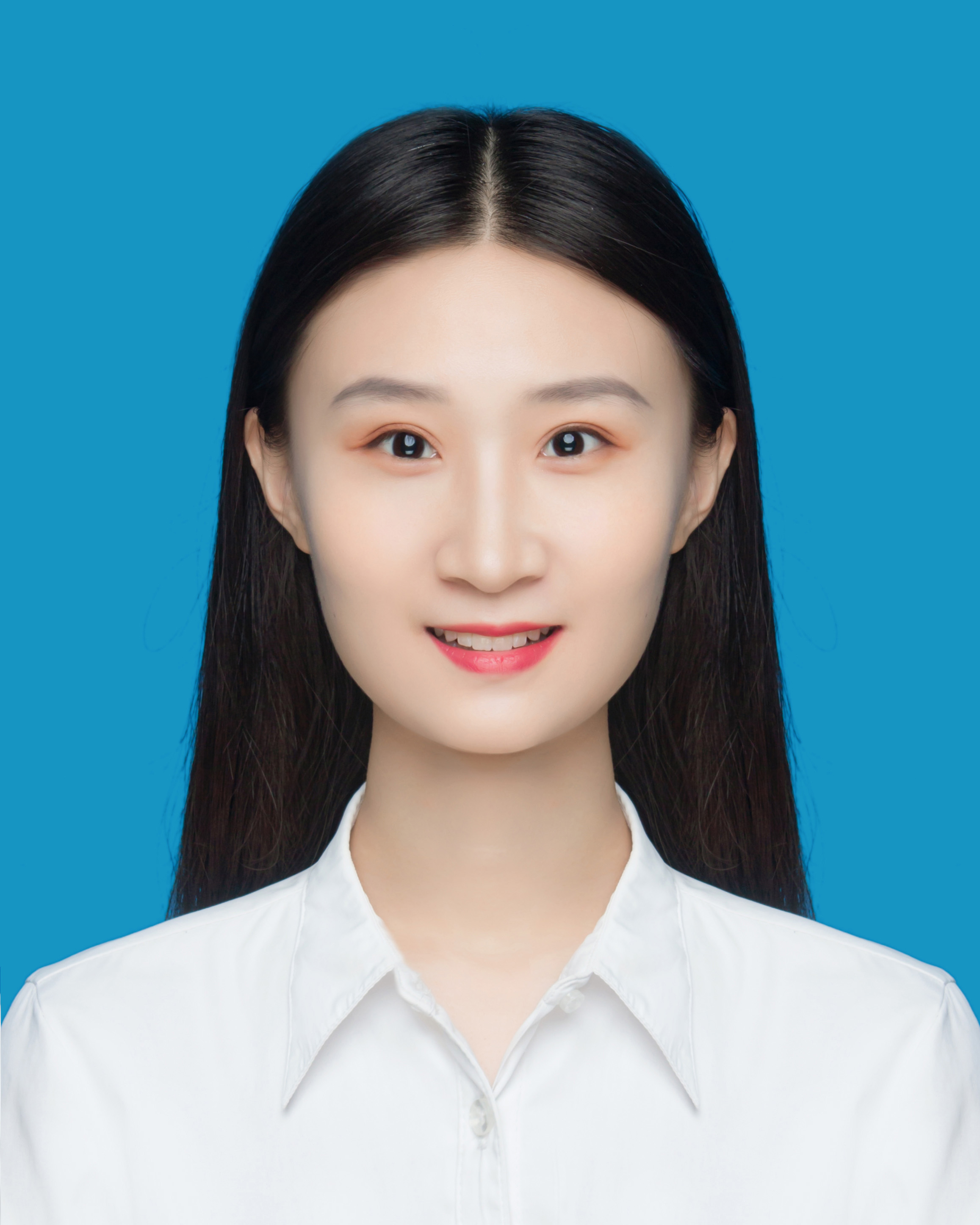}}]{Shan Zhang} received the B.Sc. degree in computer science from Shandong Agriculture University, Taian, China, in 2018, and the M.Sc. degree in compute science from Ocean University of China, Qingdao, China, in 2021.

She is currently a Assistant Researcher with the Meteorological Bureau of Tianjin Municipality. Her current research interests include computer vision and remote sensing image processing.

\end{IEEEbiography}

\begin{IEEEbiography}[{\includegraphics[width=1in,height=1.25in,clip,keepaspectratio]{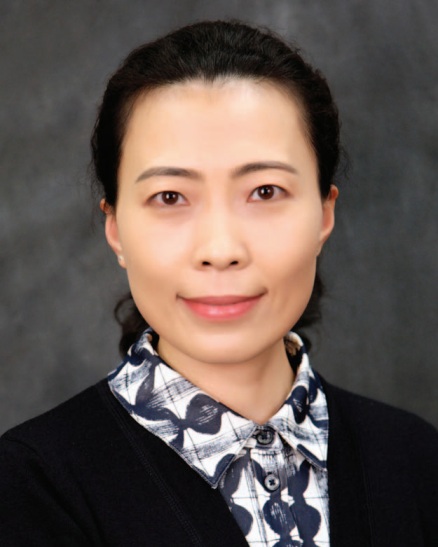}}]{Qian Du}
(Fellow, IEEE) received the Ph.D. degree in electrical engineering from the University of Maryland at Baltimore, Baltimore, MD, USA, in 2000.

She is currently the Bobby Shackouls Professor with the Department of Electrical and Computer Engineering, Mississippi State University, Starkville, MS, USA. Her research interests include hyperspectral remote-sensing image analysis and applications, and machine learning.

Dr. Du was the recipient of the 2010 Best Reviewer Award from the IEEE Geoscience and Remote Sensing Society (GRSS). She was a Co-Chair for the Data Fusion Technical Committee of the IEEE GRSS from 2009 to 2013, the Chair for the Remote Sensing and Mapping Technical Committee of International Association for Pattern Recognition from 2010 to 2014, and the General Chair for the Fourth IEEE GRSS Workshop on Hyperspectral Image and Signal Processing: Evolution in Remote Sensing held at Shanghai, China, in 2012. She was an Associate Editor
for the \textsc{IEEE Journal of Selected Topics in Applied Earth Observation and Remote Sensing}, \emph{Journal of Applied Remote Sensing}, and \textsc{IEEE Signal Processing Letters}. From
2016 to 2020, she was the Editor-in-Chief of the \textsc{IEEE Journal of Selected Topics in Applied Earth Observation and Remote Sensing}. She is currently a member of the IEEE Periodicals Review and Advisory Committee and SPIE Publications Committee. She is a Fellow of SPIE-International Society for Optics and Photonics (SPIE).

\end{IEEEbiography}

\end{document}